%% file: ScaledGD_monograph.tex
\documentclass[graybox]{svmult}
\usepackage{algorithmic}
\usepackage{algorithm}
\usepackage{amsfonts}
\usepackage{amsmath}
\usepackage{amssymb}
 
 \usepackage{geometry}
\usepackage{array}
\usepackage{arydshln}
\usepackage{bm} 
\usepackage{color}
\usepackage{comment}
\usepackage{dsfont} 
\usepackage{float}
\usepackage[T1]{fontenc}
 
\usepackage{graphicx}
\usepackage{grffile}
\usepackage{hyperref}\hypersetup{colorlinks,linkcolor=red,anchorcolor=blue,citecolor=blue}
\usepackage[latin9]{inputenc}
\usepackage{letltxmacro}
\usepackage{mathrsfs}
\usepackage{mathtools}
\usepackage{multirow}
\usepackage{nicefrac}   
\usepackage{verbatim}

\allowdisplaybreaks

\newcommand{\br}{\bm{r}}

\newcommand{\bv}{\bm{v}}

\newcommand{\bx}{\bm{x}}
\newcommand{\by}{\bm{y}}

\newcommand{\bA}{\bm{A}}

\newcommand{\bF}{\bm{F}}
\newcommand{\bG}{\bm{G}}

\newcommand{\bI}{\bm{I}}

\newcommand{\bL}{\bm{L}}
\newcommand{\bM}{\bm{M}}

\newcommand{\bQ}{\bm{Q}}
\newcommand{\bR}{\bm{R}}
\newcommand{\bS}{\bm{S}}

\newcommand{\bU}{\bm{U}}
\newcommand{\bV}{\bm{V}}
\newcommand{\bW}{\bm{W}}
\newcommand{\bX}{\bm{X}}
\newcommand{\bY}{\bm{Y}}

\newcommand{\bSigma}{\bm{\Sigma}}

\newcommand{\cA}{\mathcal{A}}

\newcommand{\cH}{\mathcal{H}}

\newcommand{\cL}{\mathcal{L}}
\newcommand{\cM}{\mathcal{M}}
\newcommand{\cN}{\mathcal{N}}

\newcommand{\cP}{\mathcal{P}}

\newcommand{\cS}{\mathcal{S}}
\newcommand{\cT}{\mathcal{T}}

\newcommand{\RR}{\mathbb{R}}

\newcommand{\bcA}{\bm{\mathcal{A}}}

\newcommand{\bcG}{\bm{\mathcal{G}}}

\newcommand{\bcS}{\bm{\mathcal{S}}}

\newcommand{\bcX}{\bm{\mathcal{X}}}
\newcommand{\bcY}{\bm{\mathcal{Y}}}

\newcommand{\argmin}{\mathop{\mathrm{argmin}}}
\newcommand{\minimize}{\mathop{\mathrm{minimize}}}
\DeclareMathOperator{\bcdot}{\boldsymbol{\cdot}}
\DeclareMathOperator{\diag}{\mathrm{diag}}
\DeclareMathOperator{\dist}{\mathrm{dist}}
\DeclareMathOperator{\fro}{\mathsf{F}}
\DeclareMathOperator{\GL}{\mathrm{GL}}

\DeclareMathOperator{\Pdiag}{\mathcal{P}_{\mathsf{diag}}}
\DeclareMathOperator{\Poffdiag}{\mathcal{P}_{\mathsf{off-diag}}}
\DeclareMathOperator{\rank}{\mathrm{rank}}
\DeclareMathOperator{\sgn}{\mathrm{sgn}}

\newcommand{\Matricize}[2]{\mathcal{M}_{#1}\left(#2\right)}
\newcommand{\Shrink}[2]{\mathcal{T}^{\mathsf{soft}}_{#1}\left[#2\right]}

\newcommand{\norm}[1]{\left\lVert#1\right\rVert}

\allowdisplaybreaks
 
\providecommand{\tabularnewline}{\\}

\let\tilde\widetilde

\usepackage{type1cm}        
\usepackage{makeidx}         
\usepackage{graphicx}        
\usepackage{multicol}        
\usepackage[bottom]{footmisc}

\usepackage{newtxtext}       %
\usepackage[varvw]{newtxmath}       

\makeindex             

\begin{document}

\title*{Provably Accelerating Ill-Conditioned Low-rank Estimation via Scaled Gradient Descent, Even with Overparameterization}
\titlerunning{Scaled Gradient Descent for Low-rank Estimation}
\author{Cong Ma, Xingyu Xu, Tian Tong, and Yuejie Chi}
\institute{Cong Ma \at University of Chicago, \email{congm@uchicago.edu}
\and Xingyu Xu \at Carnegie Mellon University, \email{xingyuxu@andrew.cmu.edu}
\and Tian Tong \at Amazon, \email{tongtn@amazon.com}
\and Yuejie Chi \at Carnegie Mellon University, \email{yuejiechi@cmu.edu}}
%
%
\maketitle

\input{abstract.tex}

\input{introduction.tex}

\input{matrix.tex}

\input{tensor.tex}

\input{overparam.tex}

\input{numerical.tex}

\input{conclusion.tex}

\begin{acknowledgement}
This work is supported in part by Office of Naval Research under N00014-19-1-2404, and by National Science Foundation under CCF-1901199, DMS-2134080 and ECCS-2126634 to Y. Chi.
\end{acknowledgement}

\bibliographystyle{plain}
\bibliography{bibfileNonconvexScaledGD,bibfileOverparam,bibfileTensor}

\end{document}

%% file: abstract.tex
\abstract*{Many problems encountered in science and engineering can be formulated as estimating a low-rank object (e.g., matrices and tensors) from incomplete, and possibly corrupted, linear measurements. Through the lens of matrix and tensor factorization, one of the most popular approaches is to employ simple iterative algorithms such as gradient descent to recover the low-rank factors directly, which allow for small memory and computation footprints. However, the convergence rate of gradient descent depends linearly, and sometimes even quadratically, on the condition number of the low-rank object, and therefore, slows down painstakingly when the problem is ill-conditioned. This chapter introduces a new algorithmic approach, dubbed scaled gradient descent (ScaledGD), that provably converges linearly at a constant rate independent of the condition number of the low-rank object, while maintaining the low per-iteration cost of gradient descent for a variety of tasks including sensing, completion and robust principal component analysis. In addition, ScaledGD continues to admit fast global convergence, again almost independent of the condition number, from a small random initialization when the rank is over-specified. In total, ScaledGD highlights the power of appropriate preconditioning in accelerating nonconvex statistical estimation, where the iteration-varying preconditioners promote desirable invariance properties of the trajectory with respect to the symmetry in low-rank factorization without hurting generalization.}

\abstract{
Many problems encountered in science and engineering can be formulated as estimating a low-rank object (e.g., matrices and tensors) from incomplete, and possibly corrupted, linear measurements. Through the lens of matrix and tensor factorization, one of the most popular approaches is to employ simple iterative algorithms such as gradient descent (GD) to recover the low-rank factors directly, which allow for small memory and computation footprints. However, the convergence rate of GD depends linearly, and sometimes even quadratically, on the condition number of the low-rank object, and therefore, GD slows down painstakingly when the problem is ill-conditioned. This chapter introduces a new algorithmic approach, dubbed scaled gradient descent (ScaledGD), that provably converges linearly at a constant rate independent of the condition number of the low-rank object, while maintaining the low per-iteration cost of gradient descent for a variety of tasks including sensing, robust principal component analysis and completion. In addition, ScaledGD continues to admit fast global convergence to the minimax-optimal solution, again almost independent of the condition number, from a small random initialization when the rank is over-specified in the presence of Gaussian noise. In total, ScaledGD highlights the power of appropriate preconditioning in accelerating nonconvex statistical estimation, where the iteration-varying preconditioners promote desirable invariance properties of the trajectory with respect to the symmetry in low-rank factorization without hurting generalization.}

%% file: introduction.tex
\section{Introduction}
\label{sec:introduction}
 Low-rank matrix and tensor estimation plays a critical role in fields such as machine learning, signal processing, imaging science, and many others. The central task can be regarded as recovering an $d$-dimensional object $\bcX_{\star} \in \RR^{n_1\times  \cdots \times n_d}$  from its highly incomplete observation $\by \in \RR^m$ given by
\begin{align*}
\by \approx  \cA(\bcX_{\star}).
\end{align*}
Here, $\cA: \RR^{n_1\times \cdots \times n_d} \mapsto \RR^m$ represents a certain linear map modeling the data collection process. Importantly, the number $m$ of observations is often much smaller than the ambient dimension $\prod_{i=1}^d n_i$ of the data object due to resource or physical constraints, necessitating the need of exploiting low-rank structures to allow for meaningful recovery. 
It is natural to minimize the least-squares loss function   
\begin{align} \label{eq:least_squares}
\minimize_{\bcX \in \RR^{n_1\times  \cdots \times n_d }}\;f(\bcX)\coloneqq \tfrac{1}{2}\|\cA(\bcX)-\by\|_{2}^{2} 
\end{align}
subject to some rank constraint. However, naively imposing the rank constraint is computationally intractable, and moreover, as the size of the object increases, the costs involved in optimizing over the full space (i.e., $\RR^{n_{1}\times \cdots \times n_{d}}$) are prohibitive in terms of both memory and computation.

To cope with these challenges, one popular approach is to represent the object of interest via its low-rank factors, which take a more economical form, and then optimize over the factors instead.
Although this leads to a nonconvex optimization problem over the factors, recent breakthroughs have shown that simple iterative algorithms such as vanilla gradient descent (GD), when properly initialized (e.g., via the spectral method), can provably converge to the true low-rank factors under mild statistical assumptions; see \cite{chi2019nonconvex} for an overview. This enables us to tap into the scalability of gradient descent in solving large-scale problems due to its amenability to computing advances such as parallelism
 \cite{bottou2018optimization}.

However, upon closer examination, the computational cost of vanilla gradient descent is still  expensive, especially for ill-conditioned objects. Although the per-iteration cost is small, the iteration complexity of gradient descent scales linearly with respect to the condition number of the low-rank matrix \cite{tu2015low}, which degenerates even worse for higher-order tensors \cite{han2020optimal}.
In fact, the issue of ill-conditioning is quite ubiquitous in real-world data modeling with many contributing factors. One one end, extracting fine-grained and weak information often manifests to estimating ill-conditioned object of interest, when the goals are to separate close-located sources of intelligence, to identify a weak mode nearby a strong one, to predict individualized responses for similar objects, and so on. While the impact of condition numbers on the computational efficacy cannot be ignored in practice, it unfortunately has not been properly addressed in recent algorithmic advances, which often assume the problem is well-conditioned.
These together raise an important question: 

\begin{svgraybox} 
Is it possible to design a first-order algorithm with a comparable per-iteration cost as gradient descent, but converges much faster at a rate that is independent of the condition number in a provable manner for a wide variety of low-rank matrix and tensor estimation tasks?
\end{svgraybox}

\subsection{An overview of ScaledGD}

In this chapter, we answer this question affirmatively by setting forth an algorithmic approach dubbed scaled gradient descent (\texttt{ScaledGD}), which is instantiated for an array of low-rank matrix and tensor estimation tasks.
 
\runinhead{ScaledGD for low-rank matrix estimation.}
By parametrizing the matrix object $\bX=\bL\bR^{\top}$ via two low-rank factors $\bL\in\RR^{n_{1}\times r}$ and $\bR\in\RR^{n_{2}\times r}$ in \eqref{eq:least_squares}, where $r$ is the rank of the true low-rank object $\bX_{\star}$, we arrive at the objective function
\begin{align}
\minimize_{\bL\in\RR^{n_1\times r},\bR\in\RR^{n_2\times r}}\;\cL(\bL,\bR)\coloneqq f(\bL\bR^{\top}).\label{eq:problem}
\end{align}
Given an initialization $(\bL_{0}, \bR_{0})$, \texttt{ScaledGD} proceeds as follows 
\begin{align}
\begin{split} \bL_{t+1} & =\bL_{t}-\eta\nabla_{\bL}\cL(\bL_{t},\bR_{t})(\bR_{t}^{\top}\bR_{t})^{-1},\\
 \bR_{t+1} & =\bR_{t}-\eta\nabla_{\bR}\cL(\bL_{t},\bR_{t})(\bL_{t}^{\top}\bL_{t})^{-1},
\end{split}\label{eq:scaledGD}
\end{align}
where $\eta > 0$ is the step size and $\nabla_{\bL}\cL(\bL_{t},\bR_{t})$ (resp.~$\nabla_{\bR}\cL(\bL_{t},\bR_{t})$) is the gradient of the loss function $\cL$ with respect to the factor $\bL_{t}$ (resp.~$\bR_{t}$) at the $t$-th iteration. Comparing to vanilla gradient descent, the search directions of the low-rank factors $\bL_{t},\bR_{t}$ in \eqref{eq:scaledGD} are {\em scaled} by $(\bR_{t}^{\top}\bR_{t})^{-1}$ and $(\bL_{t}^{\top}\bL_{t})^{-1}$ respectively. 
Intuitively, the scaling serves as a preconditioner as in quasi-Newton type algorithms, with the hope of improving the quality of the search direction to allow larger step sizes. Since computing the Hessian is extremely expensive, it is necessary to design preconditioners that are both theoretically sound and practically cheap to compute. Such requirements are met by \texttt{ScaledGD}, where the preconditioners are computed by inverting two $r\times r$ matrices, whose size is much smaller than the dimension of matrix factors.
Theoretically, we confirm that \texttt{ScaledGD} achieves linear convergence at a rate {\em independent of} the condition number of the matrix when initialized properly, e.g., using the standard spectral method, for several canonical problems: low-rank matrix sensing, robust PCA, and matrix completion.


\begin{trailer}{Performance of ScaledGD for low-rank matrix completion}
Fig.~\ref{fig:scaledGD_teaser} illustrates the relative error of completing a $1000\times 1000$ incoherent matrix (cf.~Definition~\ref{def:incoherence}) of rank $10$ with varying condition numbers from $20\%$ of its entries, using either \texttt{ScaledGD} or vanilla GD with spectral initialization.
\begin{figure}[ht]
\centering
\includegraphics[width=0.5\textwidth]{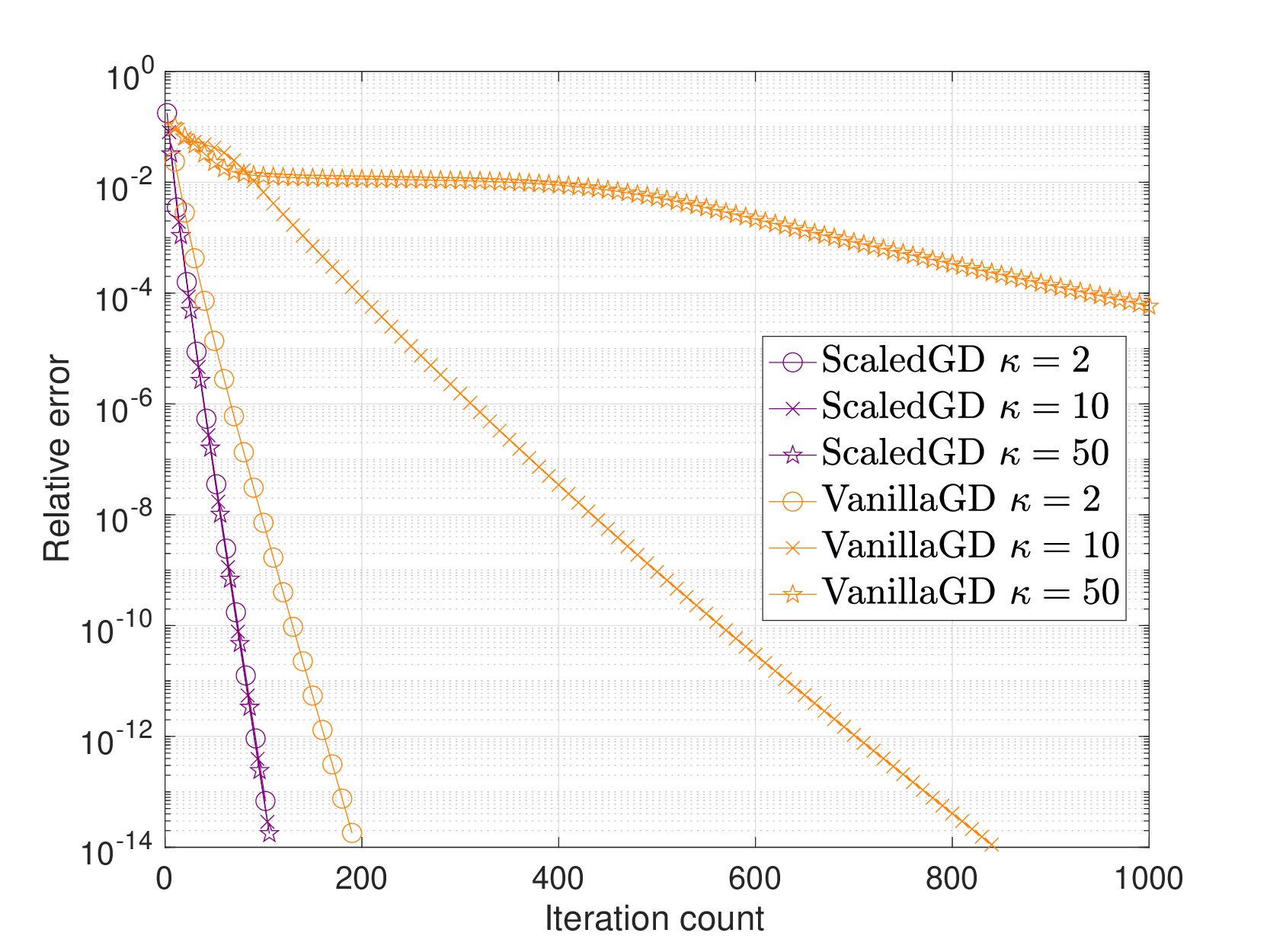}
\caption{Performance of \texttt{ScaledGD} and vanilla GD for completing a $1000\times 1000$ incoherent matrix of rank $10$ with different condition numbers $\kappa=2, 10, 50$, where each entry is observed independently with probability $0.2$. Here, both methods are initialized via the spectral method. It can be seen that \texttt{ScaledGD} converges much faster than vanilla GD even for moderately large condition numbers.}\label{fig:scaledGD_teaser}
\end{figure}
  Even for moderately ill-conditioned matrices, the convergence rate of vanilla GD slows down dramatically, while it is evident that \texttt{ScaledGD} converges at a rate independent of the condition number and therefore is much more efficient. 
\end{trailer}

\runinhead{ScaledGD for low-rank tensor estimation.} Turning to the tensor case, we focus on one of the most widely adopted low-rank structures for tensors under the {\em Tucker} decomposition \cite{tucker1966some}, by assuming the true tensor $\bcX_{\star} $ to be low-multilinear-rank, or simply low-rank, when its multilinear rank $\br=(r_1,r_2,r_3)$.  
By parameterizing the order-$3$ tensor object\footnote{For ease of presentation, we focus on 3-way tensors; our algorithm and theory can be generalized to higher-order tensors in a straightforward manner. } as $\bcX =(\bU,\bV,\bW)\bcdot\bcS $, where $\bF \coloneqq (\bU,\bV,\bW,\bcS)$ consists of the factors $\bU\in\RR^{n_1\times r_1}$, $\bV\in\RR^{n_2\times r_2}$, $\bW\in\RR^{n_3\times r_3}$, and $\bcS\in\RR^{r_1\times r_2\times r_3}$, we aim to optimize the objective function: 
\begin{align}\label{eq:loss}
\minimize_{\bF}\quad \cL(\bF)\coloneqq\frac{1}{2}\left\|\cA\left((\bU,\bV,\bW)\bcdot\bcS \right)-\by\right\|_{2}^2.
\end{align}
Given an initialization $\bF_0=(\bU_0,\bV_0,\bW_0,\bcS_0)$, \texttt{ScaledGD} proceeds as follows
\begin{align}
\bU_{t+1} &= \bU_{t} - \eta\nabla_{\bU}\cL(\bF_{t})\big(\breve{\bU}_t^{\top} \breve{\bU}_t \big)^{-1},  \nonumber\\
\bV_{t+1} &= \bV_{t} - \eta\nabla_{\bV}\cL(\bF_{t})\big(\breve{\bV}_t^{\top} \breve{\bV}_t \big)^{-1}, \nonumber\\
\bW_{t+1} &= \bW_{t} - \eta\nabla_{\bW}\cL(\bF_{t})\big(\breve{\bW}_t^{\top} \breve{\bW}_t \big)^{-1}, \nonumber \\
\bcS_{t+1} &= \bcS_{t} - \eta\left((\bU_{t}^{\top}\bU_{t})^{-1},(\bV_{t}^{\top}\bV_{t})^{-1},(\bW_{t}^{\top}\bW_{t})^{-1}\right)\bcdot\nabla_{\bcS}\cL(\bF_{t}), \label{eq:ScaledGD}
\end{align}
where $\nabla_{\bU}\cL(\bF)$, $\nabla_{\bV}\cL(\bF)$, $\nabla_{\bW}\cL(\bF)$, and  $\nabla_{\bcS}\cL(\bF)$ are the partial derivatives of $\cL(\bF)$ with respect to the corresponding variables, and 
\begin{align} \label{eq:breve_uvw}
\begin{split}
\breve{\bU}_{t}&\coloneqq\cM_{1}\left((\bI_{r_1},\bV_{t},\bW_{t})\bcdot\bcS_{t}\right)^{\top} = (\bW_{t}\otimes\bV_{t})\cM_{1}(\bcS_{t})^{\top}, \\
\breve{\bV}_{t}&\coloneqq\cM_{2}\left((\bU_{t},\bI_{r_2},\bW_{t})\bcdot\bcS_{t}\right)^{\top}=(\bW_{t}\otimes\bU_{t})\cM_{2}(\bcS_{t})^{\top},\\
\breve{\bW}_{t}&\coloneqq\cM_{3}\left((\bU_{t},\bV_{t},\bI_{r_3})\bcdot\bcS_{t}\right)^{\top}=(\bV_{t}\otimes\bU_{t})\cM_{3}(\bcS_{t})^{\top}.
\end{split}
\end{align}
Here, $\cM_k(\bcS)$ is the matricization of the tensor $\bcS$ along the $k$-th mode ($k=1,2,3$), and $\otimes$ denotes the Kronecker product. 
We investigate the theoretical properties of \texttt{ScaledGD} for tensor regression, tensor robust PCA and tensor completion, which are notably more challenging than the matrix counterpart. It is demonstrated that \texttt{ScaledGD}---when initialized properly using appropriate spectral methods---again achieves linear convergence at a rate {\em independent} of the condition number of the ground truth tensor with near-optimal sample complexities. %

\begin{trailer}{ScaledGD for low-rank tensor completion}
Fig.~\ref{fig:TC_kappa} illustrates the number of iterations needed to achieve a relative error $\|\bcX-\bcX_{\star}\|_{\fro}\le 10^{-3} \|\bcX_{\star}\|_{\fro}$ for \texttt{ScaledGD} and regularized GD \cite{han2020optimal} under different condition numbers for tensor completion under the Bernoulli sampling model. 
\begin{figure}[h]
\centering
\includegraphics[width=0.5\textwidth]{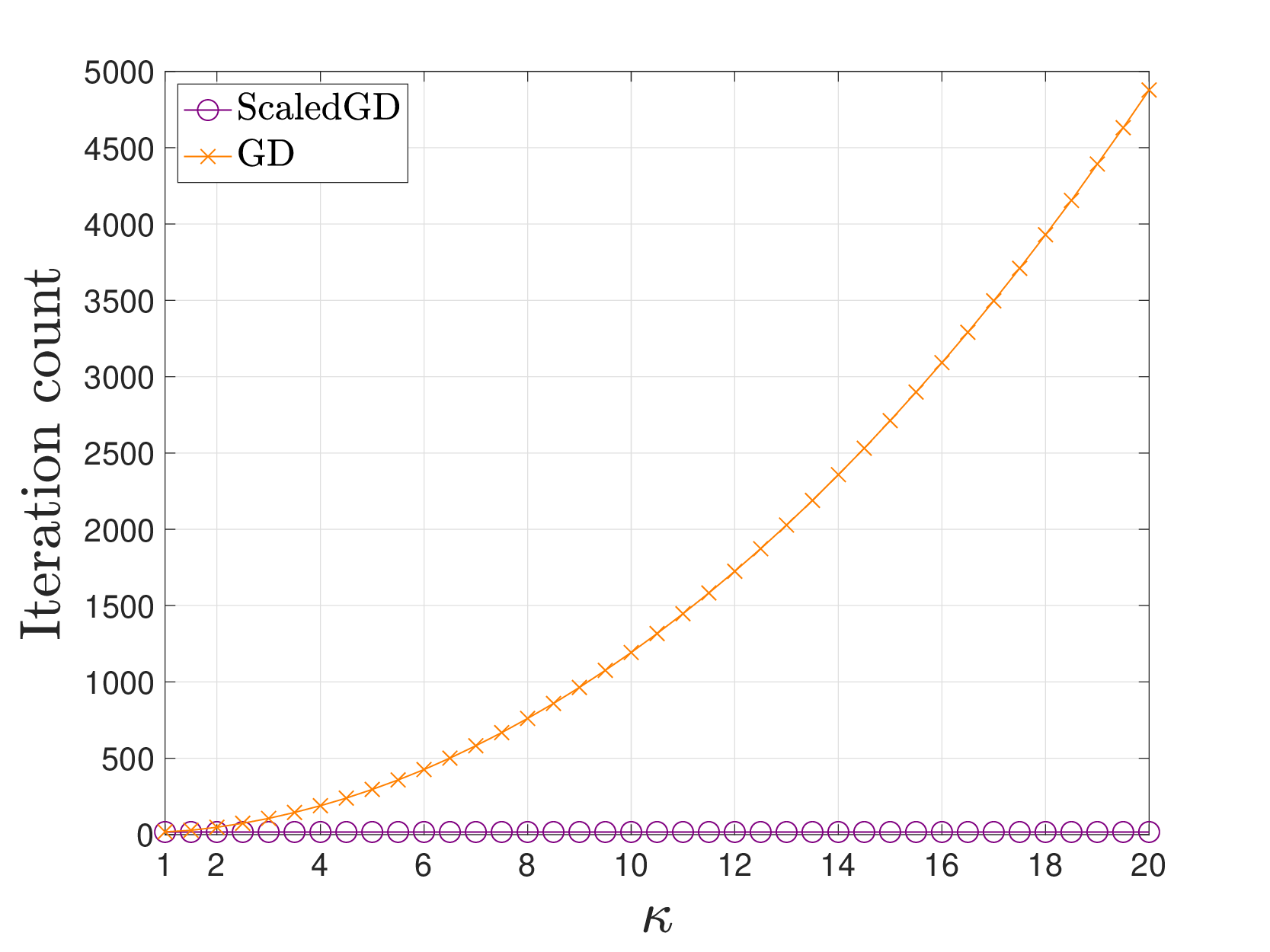} 
\caption{The iteration complexities of \texttt{ScaledGD} and regularized GD to achieve $\|\bcX-\bcX_{\star}\|_{\fro}\le 10^{-3} \|\bcX_{\star}\|_{\fro}$ with respect to different condition numbers for low-rank tensor completion with $n_1=n_2=n_3=100$, $r_1=r_2=r_3=5$, and the probability of observation $p=0.1$.}\label{fig:TC_kappa}
\end{figure} 
Clearly, the iteration complexity of GD deteriorates at a super linear rate with respect to the condition number $\kappa$, while \texttt{ScaledGD} enjoys an iteration complexity that is independent of $\kappa$ as predicted by our theory. Indeed, with a seemingly small modification, \texttt{ScaledGD} takes merely $17$ iterations to achieve the desired accuracy over the entire range of $\kappa$, while GD takes thousands of iterations even with a moderate condition number! 
\end{trailer}

To highlight, \texttt{ScaledGD} possesses many desirable properties appealing to practitioners. 
\begin{itemize}
\item {\em Low per-iteration cost:} as a preconditioned GD or quasi-Newton algorithm, \texttt{ScaledGD} updates the factors along the descent direction of a scaled gradient, where the preconditioners can be viewed as the inverse of the diagonal blocks of the Hessian for the population loss (i.e.,~matrix factorization and tensor factorization). As the sizes of the preconditioners are proportional to the rank rather than the ambient dimension, the matrix inverses are cheap to compute with a minimal overhead and the overall per-iteration cost is still low and linear in the time it takes to read the input data.
\item {\em Equivariance to parameterization:} one crucial property of \texttt{ScaledGD} is that if we reparameterize the  factors by some invertible transformation, the entire trajectory will go through the same reparameterization, leading to an {\em invariant} sequence of low-rank  updates regardless of the parameterization being adopted. 

\item {\em Implicit balancing:} \texttt{ScaledGD} optimizes the natural loss function in an {\em unconstrained} manner without requiring additional regularizations or orthogonalizations used in prior literature  \cite{han2020optimal,frandsen2020optimization,kasai2016low}, and the factors stay balanced in an automatic manner as if they are implicitly regularized \cite{ma2021beyond}.
\end{itemize}
 In total, the fast convergence rate of \texttt{ScaledGD}, together with its low computational and memory costs by operating in the factor space, makes it a highly scalable and desirable method for low-rank estimation tasks.   

\subsection{Related works}

Our work contributes to the growing literature of design and analysis of provable nonconvex optimization procedures for high-dimensional signal estimation; see e.g.~\cite{jain2017non,chen2018harnessing,chi2019nonconvex} for recent overviews.
A growing number of problems have been demonstrated to possess benign geometry that is amenable for optimization \cite{mei2016landscape} either globally or locally under appropriate statistical models. On one end, it is shown that there are no spurious local minima in the optimization landscape of matrix sensing and completion \cite{ge2016matrix,bhojanapalli2016global,park2017non,ge2017no}, phase retrieval \cite{sun2018geometric,davis2017nonsmooth}, dictionary learning \cite{sun2015complete}, kernel PCA \cite{chen2019model} and linear neural networks \cite{baldi1989neural,kawaguchi2016deep}.
Such landscape analysis facilitates the adoption of generic saddle-point escaping algorithms \cite{nesterov2006cubic,ge2015escaping,jin2017escape} to ensure global convergence. However, the resulting iteration complexity is typically high. On the other end, local refinements with carefully-designed initializations often admit fast convergence, for example in phase retrieval \cite{candes2015phase,ma2017implicit}, matrix sensing \cite{jain2013low,zheng2015convergent,wei2016guarantees}, matrix completion \cite{sun2016guaranteed,chen2015fast,ma2017implicit,chen2019nonconvex,zheng2016convergence,chen2019noisy}, blind deconvolution \cite{li2019rapid,ma2017implicit,shi2021manifold}, quadratic sampling \cite{li2021nonconvex}, and robust PCA \cite{netrapalli2014non,yi2016fast,chen2021bridging}, to name a few. 

Existing approaches for asymmetric low-rank matrix estimation often require additional regularization terms to balance the two factors, either in the form of $\frac{1}{2}\|\bL^{\top}\bL-\bR^{\top}\bR\|_{\fro}^{2}$ \cite{tu2015low,park2017non} or $\frac{1}{2}\|\bL\|_{\fro}^{2}+\frac{1}{2}\|\bR\|_{\fro}^{2}$ \cite{zhu2017global,chen2019noisy, chen2021bridging}, which ease the theoretical analysis but are often unnecessary for the practical success, as long as the initialization is balanced. Some recent work studies the unregularized gradient descent for low-rank matrix factorization and sensing including \cite{charisopoulos2019low,du2018algorithmic,ma2021beyond}.
However, the iteration complexity of all these approaches scales at least linearly with respect to the condition number $\kappa$ of the low-rank matrix, e.g.~$O(\kappa\log(1/\epsilon))$, to reach $\epsilon$-accuracy, therefore they converge slowly when the underlying matrix becomes ill-conditioned.
In contrast, \texttt{ScaledGD} enjoys a local convergence rate of $O(\log(1/\epsilon))$, therefore incurring a much smaller computational footprint when $\kappa$ is large. Last but not least, alternating minimization \cite{jain2013low,hardt2014fast} (which alternatively updates $\bL_{t}$ and $\bR_{t}$) or singular value projection \cite{netrapalli2014non,jain2010guaranteed} (which operates in the matrix space) also converge at the rate $O(\log(1/\epsilon))$, but the per-iteration cost is much higher than \texttt{ScaledGD}. Another notable algorithm is the Riemannian gradient descent algorithm in \cite{wei2016guarantees}, which also converges at the rate $O(\log(1/\epsilon))$ under the same sample complexity for low-rank matrix sensing, but requires a higher memory complexity since it operates in the matrix space rather than the factor space.

Turning to the tensor case, unfolding-based approaches typically result in sub-optimal sample complexities since they do not fully exploit the tensor structure. \cite{yuan2016tensor} studied directly minimizing the nuclear norm of the tensor, which regrettably is not computationally tractable. \cite{xia2019polynomial} proposed a Grassmannian gradient descent algorithm over the factors other than the core tensor for exact tensor completion, whose iteration complexity is not characterized. The statistical rates of tensor completion, together with a spectral method, were investigated in \cite{zhang2018tensor,xia2021statistically}, and uncertainty quantifications were recently dealt with in \cite{xia2020inference}. In addition, for low-rank tensor regression, \cite{raskutti2019convex} proposed a general convex optimization approach based on decomposable regularizers, and \cite{rauhut2017low} developed an iterative hard thresholding algorithm. A concurrent work \cite{zhang2021low} proposed a Riemannian Gauss-Newton algorithm, and obtained an impressive quadratic convergence rate for tensor regression. Compared with \texttt{ScaledGD}, this algorithm runs in the tensor space, and the update rule is more sophisticated with higher per-iteration cost by solving a least-squares problem and performing a truncated HOSVD every iteration. Another recent work \cite{cai2021generalized} studies the Riemannian gradient descent algorithm which also achieves an iteration complexity free of condition number, however, the initialization scheme was not studied therein. Riemmannian gradient descent is also applied to low-rank tensor completion with Tucker decomposition in \cite{wang2021implicit}.

\subsection{Chapter organization and notation}

The rest of this chapter is organized as follows. Section~\ref{sec:matrix} and Section~\ref{sec:tensor} describe \texttt{ScaledGD} and details its application to sensing, robust PCA and completion with theoretical guarantees in terms of both statistical and computational complexities for the matrix and tensor case respectively. Section~\ref{sec:overparam} discusses a variant of \texttt{ScaledGD} when the rank is not specified exactly. Section~\ref{sec:numerical} illustrates the empirical performance of \texttt{ScaledGD} on real data, with a particular focus on the issues of rank selection. Finally, we conclude in Section~\ref{sec:conclusion}. 

Before continuing, we introduce several notation used throughout the chapter. First of all, we use boldfaced symbols for vectors and matrices (e.g.~$\bx$ and $\bX$), and  boldface calligraphic letters (e.g.~$\bcX$) to denote tensors. For a vector $\bv$, we use $\|\bv\|_{0}$ to denote its $\ell_0$ counting norm, and $\|\bv\|_{2}$ to denote the $\ell_2$ norm. For any matrix $\bA$, we use $\sigma_{i}(\bA)$ to denote its $i$-th largest singular value, and $\sigma_{\max}(\bA)$ (resp.~$\sigma_{\min}(\bA)$) to denote its largest (resp.~smallest) nonzero singular value. Let $\Pdiag(\bA)$ denote the projection that keeps only the diagonal entries of $\bA$, and $\Poffdiag(\bA)=\bA-\Pdiag(\bA)$, for a square matrix $\bA$. Let $\bA_{i,\cdot}$ and $\bA_{\cdot,j}$ denote its $i$-th row and $j$-th column, respectively. In addition, $\|\bA\|_{\fro}$, $\|\bA\|_{1,\infty}$, $\|\bA\|_{2,\infty}$, and $\|\bA\|_{\infty}$ stand for the Frobenius norm, the $\ell_{1,\infty}$ norm (i.e.~the largest $\ell_1$ norm of the rows), the $\ell_{2,\infty}$ norm (i.e.~the largest $\ell_2$ norm of the rows), and the entrywise $\ell_{\infty}$ norm (the largest magnitude of all entries) of a matrix $\bA$. The set of invertible matrices in $\RR^{r\times r}$ is denoted by $\GL(r)$. The $r\times r$ identity matrix is denoted by $\bI_{r}$.  



The mode-$1$ matricization $\cM_1(\bcX) \in \RR^{n_1\times (n_2n_3)}$ of a tensor $\bcX \in \RR^{n_1\times n_2\times n_3}$ is given by 
$[\cM_1(\bcX)]\big(i_1, i_2 + (i_3-1)n_2\big) = \bcX(i_1,i_2,i_3)$, for $1\le i_k \le n_k$, $k=1,2,3$; $\cM_2(\bcX)$ and $\cM_3(\bcX)$ can be defined in a similar manner.
 The inner product between two tensors is defined as 
\begin{align*}
\langle\bcX_{1},\bcX_{2}\rangle = \sum_{i_1,i_2,i_3} \bcX_{1} (i_1,i_2,i_3) \bcX_{2} (i_1,i_2,i_3),
\end{align*}
and the Frobenius norm of a tensor is defined as $\|\bcX\|_{\fro}=\sqrt{\langle\bcX,\bcX\rangle}$. 
Define the $\ell_{\infty}$ norm of $\bcX$ as $\|\bcX\|_{\infty} = \max_{i_1,i_2,i_3}|\bcX(i_1,i_2,i_3)|$. With slight abuse of terminology, denote 
\begin{align*}
\sigma_{\max}(\bcX) = \max_{k=1,2,3} \sigma_{\max}(\cM_k(\bcX)), \quad\mbox{ and} \quad \sigma_{\min}(\bcX) = \min_{k=1,2,3} \sigma_{\min}(\cM_k(\bcX))
\end{align*} 
as the maximum and minimum nonzero singular values of $\bcX$.

For a general tensor $\bcX$, define $\cH_{\br}(\bcX)$ as the top-$\br$ higher-order SVD (HOSVD) of $\bcX$ with $\br= (r_1,r_2,r_3)$, given by 
\begin{align} \label{eq:HOSVD}
\cH_{\br}(\bcX)=(\bU,\bV,\bW)\bcdot\bcS,
\end{align}
where $\bU$ is the top-$r_1$ left singular vectors of $\cM_{1}(\bcX)$, $\bV$ is the top-$r_2$ left singular vectors of $\cM_{2}(\bcX)$, $\bW$ is the top-$r_3$ left singular vectors of $\cM_{3}(\bcX)$, and $\bcS = (\bU^{\top},\bV^{\top},\bW^{\top})\bcdot\bcX$ is the core tensor. 

Let $a\vee b=\max\{a,b\}$ and $a\wedge b=\min\{a,b\}$. Throughout, $f(n)\lesssim g(n)$ or $f(n)=O(g(n))$ means $|f(n)|/|g(n)|\le C$
for some constant $C>0$ when $n$ is sufficiently large; $f(n)\gtrsim g(n)$ means $|f(n)|/|g(n)|\ge C$
for some constant $C>0$ when $n$ is sufficiently large.
Last but not least, we use the terminology ``with overwhelming probability'' to denote the event happens with probability at least $1-c_{1}n^{-c_{2}}$, where $c_{1},c_{2}>0$ are some universal constants, whose values may vary from line to line.

%% file: matrix.tex
\section{ScaledGD for Low-Rank Matrix Estimation}
\label{sec:matrix}

This section is devoted to introducing \texttt{ScaledGD} and establishing its statistical and computational guarantees for various low-rank matrix estimation problems; the majority of the results are based on \cite{tong2021accelerating}.   Table~\ref{tab:performance-guarantees-ScaledGD} summarizes the performance guarantees of \texttt{ScaledGD} in terms of both statistical and computational complexities with comparisons to prior algorithms using GD.

\begin{table}[ht]
 \caption{Comparisons of \texttt{ScaledGD} with \texttt{GD} when tailored to various problems (with spectral initialization) \cite{tu2015low,yi2016fast,zheng2016convergence}, where they have comparable per-iteration costs. Here, we say that the output $\bX$ of an algorithm reaches $\epsilon$-accuracy, if it satisfies $\|\bX-\bX_{\star}\|_{\fro}\le\epsilon\sigma_{r}(\bX_{\star})$. Here, $n\coloneqq n_1 \vee n_2 = \max\{n_{1},n_{2}\}$, $\kappa$ and $\mu$ are the condition number and incoherence parameter of $\bX_{\star}$.
 \label{tab:performance-guarantees-ScaledGD}  }
\centering %
\begin{tabular}{c||c|c||c|c||c|c}
\hline 
 & \multicolumn{2}{c||}{Matrix sensing}   & \multicolumn{2}{c||}{Matrix robust PCA}  & \multicolumn{2}{c}{Matrix completion}  \tabularnewline
\hline 
\hline 
\multirow{2}{*}{Algorithms} & sample  & iteration & corruption  & iteration  & sample  & iteration \tabularnewline
 & complexity  & complexity     & fraction  & complexity &   complexity  & complexity \tabularnewline
\hline 
\multirow{2}{*}{\texttt{GD}}  & \multirow{2}{*}{$nr^{2}\kappa^{2}$} & \multirow{2}{*}{$\kappa\log\frac{1}{\epsilon}$} &   \multirow{2}{*}{$\frac{1}{\mu r^{3/2}\kappa^{3/2} \vee \mu r\kappa^{2}}$} & \multirow{2}{*}{$\kappa\log\frac{1}{\epsilon}$}   & \multirow{2}{*}{$(\mu\vee\log n)\mu n r^{2}\kappa^{2}$ } & \multirow{2}{*}{$\kappa\log\frac{1}{\epsilon}$} \tabularnewline
 &  &    &  &   &  &  \tabularnewline 
\hline 
\multirow{2}{*}{\texttt{ScaledGD}}  & \multirow{2}{*}{$nr^{2}\kappa^{2}$} & \multirow{2}{*}{$\log\frac{1}{\epsilon}$}   & \multirow{2}{*}{$\frac{1}{\mu r^{3/2}\kappa}$} & \multirow{2}{*}{$\log\frac{1}{\epsilon}$}    & \multirow{2}{*}{$(\mu\kappa^{2}\vee\log n)\mu n r^{2}\kappa^{2} $} & \multirow{2}{*}{$\log\frac{1}{\epsilon}$}\tabularnewline
  &  &    &  &     & & \tabularnewline
\hline
\end{tabular}
\end{table}

\subsection{Assumptions} \label{subsec:scaledGD_assumptions}

Denote by $\bU_{\star}\bSigma_{\star}\bV_{\star}^{\top}$ the compact singular value decomposition (SVD) of the rank-$r$ matrix $\bX_{\star}\in\RR^{n_{1}\times n_{2}}$, i.e.,
$$  \bX_{\star} = \bU_{\star}\bSigma_{\star}\bV_{\star}^{\top}. $$
Here, $\bU_{\star}\in\RR^{n_{1}\times r}$ and $\bV_{\star}\in\RR^{n_{2}\times r}$ are composed of $r$ left and right singular vectors, respectively, and $\bSigma_{\star}\in\RR^{r\times r}$ is a diagonal matrix consisting of $r$ singular values of $\bX_{\star}$ organized in a non-increasing order, i.e.~$\sigma_{1}(\bX_{\star})\ge\dots\ge\sigma_{r}(\bX_{\star})>0$. Define the ground truth low-rank factors as 
\begin{align}
\bL_{\star}\coloneqq\bU_{\star}\bSigma_{\star}^{1/2}, \qquad\mbox{and}\qquad \bR_{\star}\coloneqq\bV_{\star}\bSigma_{\star}^{1/2},\label{eq:true_SVD}
\end{align} 
so that $\bX_{\star}=\bL_{\star}\bR_{\star}^{\top}$. Correspondingly, denote the stacked factor matrix as
\begin{align}
\bF_{\star}\coloneqq\begin{bmatrix}\bL_{\star} \\ \bR_{\star}\end{bmatrix}\in\RR^{(n_{1}+n_{2})\times r}.\label{eq:true_factor}
\end{align} 

\runinhead{Key parameters.} The condition number of $\bX_{\star}$ is defined as follows.
\begin{definition}[Matrix condition number]Define 
\begin{align}
\kappa\coloneqq \frac{\sigma_{1}(\bX_{\star})}{ \sigma_{r}(\bX_{\star})}\label{eq:kappa}
\end{align} 
as the condition number of $\bX_{\star}$. 
\end{definition}

We next introduce the incoherence condition, which is known to be crucial for reliable estimation of the low-rank matrix $\bX_{\star}$ in matrix completion and robust PCA~\cite{chen2015incoherence}. 
\begin{definition}[Matrix incoherence]\label{def:incoherence} A rank-$r$ matrix $\bX_{\star}\in\RR^{n_{1}\times n_{2}}$ with compact SVD as $\bX_{\star}=\bU_{\star}\bSigma_{\star}\bV_{\star}^{\top}$ is said to be $\mu$-incoherent if
\begin{align*}
\|\bU_{\star}\|_{2,\infty}\le\sqrt{\frac{\mu}{n_{1}} }\|\bU_{\star}\|_{\fro}=\sqrt{\frac{\mu r}{n_{1}}},\quad\mbox{and}\quad\|\bV_{\star}\|_{2,\infty}\le\sqrt{\frac{\mu}{n_{2}}}\|\bV_{\star}\|_{\fro}=\sqrt{ \frac{\mu r}{n_{2}}}.
\end{align*}
\end{definition}

\subsection{Matrix sensing}\label{subsec:scaledGD_MS}

\runinhead{Observation model.} Assume that we have collected a set of linear measurements about a rank-$r$ matrix $\bX_{\star}\in\RR^{n_{1}\times n_{2}}$, given as 
\begin{align}
\by=\cA(\bX_{\star})\in\RR^{m},\label{eq:sensing_measurements}
\end{align}
where $\cA(\bX)=\{\langle\bA_{i},\bX\rangle\}_{i=1}^{m}:\RR^{n_{1}\times n_{2}}\mapsto\RR^{m}$ is the linear map modeling the measurement process. The goal of low-rank matrix sensing is to recover $\bX_{\star}$ from $\by$ when the number of measurements $m\ll n_{1}n_{2}$, which 
has wide applications in medical imaging, signal processing, and data compression \cite{candes2011tight}. 

\runinhead{Algorithm development.} Writing $\bX\in\RR^{n_{1}\times n_{2}}$ into a factored form $\bX=\bL\bR^{\top}$, we consider the following optimization problem: 
\begin{align}
\minimize_{\bF\in\RR^{(n_{1}+n_{2})\times r}}\;\cL(\bF)=\frac{1}{2}\left\Vert \cA(\bL\bR^{\top})-\by\right\Vert _{2}^{2}.\label{eq:loss_MS}
\end{align}
Here as before, $\bF$ denotes the stacked factor matrix $[\bL^\top, \bR^\top]^\top$. We suggest running \texttt{ScaledGD} \eqref{eq:scaledGD} with the spectral initialization to solve \eqref{eq:loss_MS}, which performs the top-$r$ SVD on $\cA^{*}(\by)$, where $\cA^{*}(\cdot)$ is the adjoint operator of $\cA(\cdot)$. The full algorithm is stated in Algorithm~\ref{alg:MS}. The low-rank matrix can be estimated as $\bX_{T}=\bL_{T}\bR_{T}^{\top}$ after running $T$ iterations of \texttt{ScaledGD}.

\begin{algorithm}[t]
\caption{\texttt{ScaledGD} for low-rank matrix sensing with spectral initialization}\label{alg:MS} 
\begin{algorithmic} \STATE \textbf{{Spectral initialization}}: Let $\bU_{0}\bSigma_{0}\bV_{0}^{\top}$ be the top-$r$ SVD of $\cA^{*}(\by)$, and set 
\begin{align}
\bL_{0}=\bU_{0}\bSigma_{0}^{1/2},\quad\mbox{and}\quad\bR_{0}=\bV_{0}\bSigma_{0}^{1/2}.\label{eq:init_MS}
\end{align}
\vspace{-0.1in}
\STATE \textbf{{Scaled gradient updates}}: \textbf{for} $t=0,1,2,\dots,T-1$ \textbf{do} 
\begin{align}
\begin{split} & \bL_{t+1}=\bL_{t}-\eta\cA^{*}(\cA(\bL_{t}\bR_{t}^{\top})-\by)\bR_{t}(\bR_{t}^{\top}\bR_{t})^{-1},\\
 & \bR_{t+1}=\bR_{t}-\eta\cA^{*}(\cA(\bL_{t}\bR_{t}^{\top})-\by)^{\top}\bL_{t}(\bL_{t}^{\top}\bL_{t})^{-1}.
\end{split}\label{eq:iterates_MS}
\end{align}
\end{algorithmic} 
\end{algorithm}

\runinhead{Theoretical guarantee.} To understand the performance of \texttt{ScaledGD} for low-rank matrix sensing, we adopt a standard assumption on the sensing operator $\cA(\cdot)$, namely the Restricted Isometry Property (RIP). 
\begin{definition}[Matrix RIP \cite{recht2010guaranteed}] The linear map $\cA(\cdot)$ is said to obey the rank-$r$ RIP with a constant $\delta_{r}\in[0,1)$, if for all matrices $\bM\in\RR^{n_{1}\times n_{2}}$ of rank at most $r$, one has 
\begin{align*}
(1-\delta_{r})\|\bM\|_{\fro}^{2}\le\|\cA(\bM)\|_{2}^{2}\le(1+\delta_{r})\|\bM\|_{\fro}^{2}.
\end{align*}
\end{definition}

It is well-known that many measurement ensembles satisfy the RIP property \cite{recht2010guaranteed,candes2011tight}. For example, if the entries of $\bA_{i}$'s are composed of i.i.d.~Gaussian entries $\cN(0,1/m)$, then the RIP is satisfied for a constant $\delta_{r}$ as long as $m$ is on the order of $(n_{1}+n_{2})r/\delta_{r}^{2}$. With the RIP condition in place, the following theorem demonstrates that \texttt{ScaledGD} converges linearly at a constant rate as long as the sensing operator $\cA(\cdot)$ has a sufficiently small RIP constant.

\begin{theorem}\label{thm:MS} Suppose that $\cA(\cdot)$ obeys the $2r$-RIP with $\delta_{2r}\le0.02/(\sqrt{r}\kappa)$. If the step size obeys $0<\eta\le2/3$, then for all $t\ge0$, the iterates of the \texttt{ScaledGD} method in Algorithm~\ref{alg:MS} satisfy 
\begin{align*}
\left\Vert \bL_{t}\bR_{t}^{\top}-\bX_{\star}\right\Vert _{\fro}& \le(1-0.6\eta)^{t}0.15\sigma_{r}(\bX_{\star}).
\end{align*}
\end{theorem}

Theorem~\ref{thm:MS} establishes that the reconstruction error $\|\bL_{t}\bR_{t}^{\top}-\bX_{\star}\|_{\fro}$ contracts linearly at a constant rate, as long as the sample size satisfies $m=O(nr^{2}\kappa^{2})$ with Gaussian random measurements \cite{recht2010guaranteed}, where we recall that $n=n_{1}\vee n_{2}$.
To reach $\epsilon$-accuracy, i.e.~$\|\bL_{t}\bR_{t}^{\top}-\bX_{\star}\|_{\fro}\le\epsilon\sigma_{r}(\bX_{\star})$, \texttt{ScaledGD} takes at most $T=O(\log(1/\epsilon))$ iterations, which is {\em independent} of the condition number $\kappa$ of $\bX_{\star}$. 
In comparison, GD with spectral initialization in \cite{tu2015low} converges in $O(\kappa\log(1/\epsilon))$ iterations as long as $m=O(nr^{2}\kappa^{2})$.
Therefore, \texttt{ScaledGD} converges at a much faster rate than GD at the same sample complexity while maintaining a similar per-iteration cost (cf.~Table~\ref{tab:performance-guarantees-ScaledGD}).

\subsection{Matrix robust principal component analysis}\label{subsec:scaledGD_RPCA}

\runinhead{Observation model.} Assume that we have observed the data matrix 
\begin{align*}
\bY=\bX_{\star}+\bS_{\star},
\end{align*}
which is a superposition of a rank-$r$ matrix $\bX_{\star}$, modeling the clean data, and a sparse matrix $\bS_{\star}$, modeling the corruption or outliers.
The goal of robust PCA~\cite{candes2009robustPCA,chandrasekaran2011siam} is to separate the two matrices $\bX_{\star}$ and $\bS_{\star}$ from their mixture $\bY$.  

Following~\cite{chandrasekaran2011siam,netrapalli2014non,yi2016fast}, we consider a deterministic sparsity model for $\bS_{\star}$, in which $\bS_{\star}$ contains at most $\alpha$-fraction of nonzero entries per row and column for some $\alpha\in[0,1)$, i.e.~$\bS_{\star}\in\cS_{\alpha}$, where we denote
\begin{align}
\cS_{\alpha}\coloneqq\{\bS\in\RR^{n_{1}\times n_{2}}:\|\bS_{i,\cdot}\|_{0}\le\alpha n_{2}\mbox{ for all }i, \mbox{ and }\|\bS_{\cdot,j}\|_{0}\le\alpha n_{1}\mbox{ for all }j\}.\label{eq:S_alpha}
\end{align}

\runinhead{Algorithm development.} Writing $\bX\in\RR^{n_{1}\times n_{2}}$ into the factored form $\bX=\bL\bR^{\top}$, we consider the following optimization problem: 
\begin{align}
\minimize_{\bF\in\RR^{(n_{1}+n_{2})\times r},\bS\in\cS_{\alpha}}\;\cL(\bF,\bS)=\frac{1}{2}\left\Vert \bL\bR^{\top}+\bS-\bY\right\Vert _{\fro}^{2}.\label{eq:loss_RPCA}
\end{align}
It is thus natural to alternatively update $\bF=[\bL^{\top},\bR^{\top}]^{\top}$ and $\bS$, where $\bF$ is updated via the proposed \texttt{ScaledGD} algorithm, and $\bS$ is updated by hard thresholding, which trims the small entries of the residual matrix $\bY-\bL\bR^{\top}$. More specifically, for some truncation level $0\le\bar{\alpha}\le1$, we define the sparsification operator that only keeps $\bar{\alpha}$ fraction of largest entries in each row and column:
\begin{align}
(\cT_{\bar{\alpha}}[\bA])_{i,j}=\begin{cases}
\bA_{i,j}, & \mbox{if }|\bA|_{i,j}\ge|\bA|_{i,(\bar{\alpha}n_{2})},\mbox{ and }|\bA|_{i,j}\ge|\bA|_{(\bar{\alpha}n_{1}),j}\\
0, & \mbox{otherwise}
\end{cases},\label{eq:T_alpha}
\end{align}
where $|\bA|_{i,(k)}$ (resp.~$|\bA|_{(k),j}$) denote the $k$-th largest element in magnitude in the $i$-th row (resp.~$j$-th column).
The \texttt{ScaledGD} algorithm with the spectral initialization for solving robust PCA is formally stated in Algorithm~\ref{alg:RPCA}. Note that, comparing with \cite{yi2016fast}, we do not require a balancing term $\|\bL^{\top}\bL-\bR^{\top}\bR\|_{\fro}^{2}$ in the loss function \eqref{eq:loss_RPCA}, nor the projection of the low-rank factors onto the $\ell_{2,\infty}$ ball in each iteration.
\begin{algorithm}[ht]
\caption{\texttt{ScaledGD} for robust PCA with spectral initialization}\label{alg:RPCA} 
\begin{algorithmic} \STATE \textbf{{Spectral initialization}}: Let $\bU_{0}\bSigma_{0}\bV_{0}^{\top}$ be the top-$r$ SVD of $\bY-\cT_{\alpha}[\bY]$, and set 
\begin{align}
\bL_{0}=\bU_{0}\bSigma_{0}^{1/2},\quad\mbox{and}\quad\bR_{0}=\bV_{0}\bSigma_{0}^{1/2}.\label{eq:init_RPCA}
\end{align} 
\vspace{-0.1in}
\STATE \textbf{{Scaled gradient updates}}: \textbf{for} $t=0,1,2,\dots,T-1$ \textbf{do} 
\begin{align}
\begin{split} \bS_{t} & =\cT_{2\alpha}[\bY-\bL_{t}\bR_{t}^{\top}],\\
 \bL_{t+1} & =\bL_{t}-\eta(\bL_{t}\bR_{t}^{\top}+\bS_{t}-\bY)\bR_{t}(\bR_{t}^{\top}\bR_{t})^{-1},\\
 \bR_{t+1} & =\bR_{t}-\eta(\bL_{t}\bR_{t}^{\top}+\bS_{t}-\bY)^{\top}\bL_{t}(\bL_{t}^{\top}\bL_{t})^{-1}.
\end{split}\label{eq:iterates_RPCA}
\end{align}
\end{algorithmic} 
\end{algorithm}

\runinhead{Theoretical guarantee.} 
The following theorem establishes the performance guarantee of \texttt{ScaledGD} as long as the fraction $\alpha$ of corruptions is sufficiently small.

\begin{theorem}\label{thm:RPCA} Suppose that $\bX_{\star}$ is $\mu$-incoherent and that the corruption fraction $\alpha$ obeys $\alpha\le c/(\mu r^{3/2}\kappa)$ for some sufficiently small constant $c>0$. If the step size obeys $0.1\le\eta\le2/3$, then for all $t\ge0$, the iterates of \texttt{ScaledGD} in Algorithm~\ref{alg:RPCA} satisfy 
\begin{align*}
\left\Vert \bL_{t}\bR_{t}^{\top}-\bX_{\star}\right\Vert _{\fro} & \le(1-0.6\eta)^{t}0.03\sigma_{r}(\bX_{\star}).
\end{align*}
\end{theorem}

Theorem~\ref{thm:RPCA} establishes that the reconstruction error $\| \bL_{t}\bR_{t}^{\top}-\bX_{\star}\|_{\fro}$ contracts linearly at a constant rate, as long as the fraction of corruptions satisfies $\alpha\lesssim1/(\mu r^{3/2}\kappa)$.
To reach $\epsilon$-accuracy, i.e.~$\|\bL_{t}\bR_{t}^{\top}-\bX_{\star}\|_{\fro}\le\epsilon\sigma_{r}(\bX_{\star})$, \texttt{ScaledGD} takes at most $T=O(\log(1/\epsilon))$ iterations, which is {\em independent} of $\kappa$. 
In comparison, projected gradient descent with spectral initialization in \cite{yi2016fast} converges in $O(\kappa\log(1/\epsilon))$ iterations as long as $\alpha\lesssim 1/(\mu r^{3/2}\kappa^{3/2}\vee\mu r\kappa^{2})$.
Therefore, \texttt{ScaledGD} converges at a much faster rate than GD while maintaining a comparable per-iteration cost (cf.~Table~\ref{tab:performance-guarantees-ScaledGD}). In addition, our theory unveils that \texttt{ScaledGD} automatically maintains the incoherence and balancedness of the low-rank factors without imposing explicit regularizations.

\subsection{Matrix completion}\label{subsec:scaledGD_MC}

\runinhead{Observation model.} Assume that we have observed a subset $\Omega$ of entries of $\bX_{\star}$ given as 
$\cP_{\Omega}(\bX_{\star})$, where $\cP_{\Omega}:\RR^{n_{1}\times n_{2}}\mapsto\RR^{n_{1}\times n_{2}}$ is a projection such that
\begin{align}
(\cP_{\Omega}(\bX))_{i,j}=\begin{cases}\bX_{i,j}, & \mbox{if }(i,j)\in\Omega, \\
0, & \mbox{otherwise}\end{cases}.
\end{align}
Here, $\Omega$ is generated according to the Bernoulli model in the sense that each $(i,j) \in \Omega$ independent with probability $p \in (0,1]$. 
The goal of matrix completion is to recover the matrix $\bX_{\star}$ from its partial observation $\cP_{\Omega}(\bX_{\star})$. 

\runinhead{Algorithm development.} Again, writing $\bX\in\RR^{n_{1}\times n_{2}}$ into the factored form $\bX=\bL\bR^{\top}$, we consider the following optimization problem: 
\begin{align}
\minimize_{\bF\in\RR^{(n_{1}+n_{2})\times r}}\;\cL(\bF)\coloneqq\frac{1}{2p}\left\Vert \cP_{\Omega}(\bL\bR^{\top}-\bX_{\star})\right\Vert _{\fro}^2.\label{eq:loss_MC}
\end{align}
Similarly to robust PCA, the underlying low-rank matrix $\bX_{\star}$ needs to be incoherent (cf.~Definition~\ref{def:incoherence}) to avoid ill-posedness. One typical strategy to ensure the incoherence condition is to perform projection after the gradient update, by projecting the iterates to maintain small $\ell_{2,\infty}$ norms of the factor matrices. However, the standard projection operator \cite{chen2015fast} is not covariant with respect to invertible transforms, and consequently, needs to be modified when using scaled gradient updates. To that end, we introduce the following new projection operator: for every $\tilde{\bF}\in\RR^{(n_{1}+n_{2})\times r} = [\tilde{\bL}^{\top}, \tilde{\bR}^{\top}]^{\top}$,
\begin{align}
\begin{split}
 \cP_{B}(\tilde{\bF}) =& \argmin_{\bF\in\RR^{(n_{1}+n_{2})\times r}}\; \big\|(\bL-\tilde{\bL})(\tilde{\bR}^{\top}\tilde{\bR})^{1/2}\big\|_{\fro}^{2} + \big\Vert(\bR-\tilde{\bR})(\tilde{\bL}^{\top}\tilde{\bL})^{1/2}\big\Vert_{\fro}^{2} \\
 & \quad\mbox{s.t.}\quad\sqrt{n_{1}}\big\Vert\bL(\tilde{\bR}^{\top}\tilde{\bR})^{1/2}\big\Vert_{2,\infty} \vee \sqrt{n_{2}}\big\Vert\bR(\tilde{\bL}^{\top}\tilde{\bL})^{1/2}\big\Vert_{2,\infty} \le B
\end{split},\label{eq:scaled_proj_opt}
\end{align} 
which finds a factored matrix that is closest to $\tilde{\bF}$ and stays incoherent in a weighted sense. Luckily, the solution to the above scaled projection admits a simple closed-form solution, given by
\begin{align}
\begin{split}
 \cP_{B}(\tilde{\bF}) \coloneqq\begin{bmatrix}\bL \\ \bR \end{bmatrix},\quad\mbox{where}\,\,
\bL_{i,\cdot} & \coloneqq \left(1\wedge \frac{B}{\sqrt{n_{1}}\|\tilde{\bL}_{i,\cdot}\tilde{\bR}^{\top}\|_{2}}\right)\tilde{\bL}_{i,\cdot}, \, 1\le i \le n_{1}, \\ 
\bR_{j,\cdot} & \coloneqq \left(1\wedge \frac{B}{\sqrt{n_{2}}\|\tilde{\bR}_{j,\cdot}\tilde{\bL}^{\top}\|_{2}}\right)\tilde{\bR}_{j,\cdot}, \, 1\le j \le n_{2}.
\end{split}\label{eq:scaled_proj}
\end{align}
 
With the new projection operator in place, we propose the following \texttt{ScaledGD} method with spectral initialization for solving matrix completion, formally stated in Algorithm~\ref{alg:MC}.

\begin{algorithm}[ht]
\caption{\texttt{ScaledGD} for matrix completion with spectral initialization}\label{alg:MC} 
\begin{algorithmic} \STATE \textbf{{Spectral initialization}}: Let $\bU_{0}\bSigma_{0}\bV_{0}^{\top}$ be the top-$r$ SVD of $\frac{1}{p}\cP_{\Omega}(\bX_{\star})$, and set
\begin{align}
\begin{bmatrix}\bL_{0} \\ \bR_{0}\end{bmatrix}=\cP_{B}\left(\begin{bmatrix}\bU_{0}\bSigma_{0}^{1/2} \\ \bV_{0}\bSigma_{0}^{1/2}\end{bmatrix}\right).\label{eq:init_MC}
\end{align}
\vspace{-0.1in}
\STATE \textbf{{Scaled projected gradient updates}}: \textbf{for} $t=0,1,2,\dots,T-1$ \textbf{do} 
\begin{align}
\begin{bmatrix}\bL_{t+1} \\ \bR_{t+1}\end{bmatrix}=\cP_{B}\left(\begin{bmatrix}
\bL_{t}-\frac{\eta}{p}\cP_{\Omega}(\bL_{t}\bR_{t}^{\top}-\bX_{\star})\bR_{t}(\bR_{t}^{\top}\bR_{t})^{-1} \\
\bR_{t}-\frac{\eta}{p}\cP_{\Omega}(\bL_{t}\bR_{t}^{\top}-\bX_{\star})^{\top}\bL_{t}(\bL_{t}^{\top}\bL_{t})^{-1}
\end{bmatrix}\right).\label{eq:iterates_MC}
\end{align}
\end{algorithmic} 
\end{algorithm}

\runinhead{Theoretical guarantee.}   The following theorem establishes the performance guarantee of \texttt{ScaledPGD} as long as the number of observations is sufficiently large.
 
\begin{theorem}\label{thm:MC} Suppose that $\bX_{\star}$ is $\mu$-incoherent, and that $p$ satisfies $p\ge C(\mu\kappa^{2}\vee \log(n_{1}\vee n_{2}))\mu r^{2}\kappa^{2}/(n_{1}\wedge n_{2})$ for some sufficiently large constant $C$. Set the projection radius as $B=C_{B}\sqrt{\mu r}\sigma_{1}(\bX_{\star})$ for some constant $C_{B}\ge1.02$. If the step size obeys $0<\eta\le2/3$, then with probability at least $1-c_{1}(n_{1}\vee n_{2})^{-c_{2}}$, for all $t\ge0$, the iterates of \texttt{ScaledGD} in \eqref{eq:iterates_MC} satisfy
\begin{align*}
\left\|\bL_{t}\bR_{t}^{\top}-\bX_{\star}\right\|_{\fro} &\le(1-0.6\eta)^{t}0.03\sigma_{r}(\bX_{\star}).
\end{align*} 
Here $c_{1},c_{2}>0$ are two universal constants. 
\end{theorem}

Theorem~\ref{thm:MC} establishes that the reconstruction error $\|\bL_{t}\bR_{t}^{\top}-\bX_{\star}\|_{\fro}$ contracts linearly at a constant rate, as long as the probability of observation satisfies $p \gtrsim (\mu\kappa^{2}\vee \log(n_{1}\vee n_{2}))\mu r^{2}\kappa^{2}/(n_{1}\wedge n_{2})$.
To reach $\epsilon$-accuracy, i.e.~$\|\bL_{t}\bR_{t}^{\top}-\bX_{\star}\|_{\fro}\le\epsilon\sigma_{r}(\bX_{\star})$, \texttt{ScaledPGD} takes at most $T=O(\log(1/\epsilon))$ iterations, which is {\em independent} of $\kappa$. In comparison, projected gradient descent \cite{zheng2016convergence} with spectral initialization converges in $O(\kappa\log(1/\epsilon))$ iterations as long as $p \gtrsim (\mu\vee\log(n_{1}\vee n_{2}))\mu r^{2}\kappa^{2}/(n_{1}\wedge n_{2})$. Therefore, \texttt{ScaledGD} achieves much faster convergence than its unscaled counterpart, at a slightly higher sample complexity, which we believe can be further improved by finer analysis (cf.~Table~\ref{tab:performance-guarantees-ScaledGD}).   

\subsection{A glimpse of the analysis}

At the heart of our analysis is a proper metric to measure the progress of the \texttt{ScaledGD} iterates $\bF_{t}\coloneqq[\bL_{t}^{\top},\bR_{t}^{\top}]^{\top}$. 
Obviously, the factored representation is not unique in that for any invertible matrix $\bQ\in\GL(r)$, one has $\bL\bR^\top=(\bL\bQ)(\bR\bQ^{-\top})^{\top}$. Therefore, the reconstruction error metric needs to take into account this identifiability issue. More importantly, we need a diagonal scaling in the distance error metric to properly account for the effect of preconditioning. To provide intuition, note that the update rule \eqref{eq:scaledGD} can be viewed as finding the best local quadratic approximation of $\cL(\cdot)$ in the following sense:
\begin{align*}
& (\bL_{t+1},\bR_{t+1}) \\
& =\argmin_{\bL,\bR}\;  \cL(\bL_{t},\bR_{t})+\langle\nabla_{\bL}\cL(\bL_{t},\bR_{t}),\bL-\bL_{t}\rangle+\langle\nabla_{\bR}\cL(\bL_{t},\bR_{t}),\bR-\bR_{t}\rangle \\ 
 &\qquad\qquad\qquad +\frac{1}{2\eta}\left(\left\Vert (\bL-\bL_{t})(\bR_{t}^{\top}\bR_{t})^{1/2}\right\Vert _{\fro}^{2}+\left\Vert (\bR-\bR_{t})(\bL_{t}^{\top}\bL_{t})^{1/2}\right\Vert _{\fro}^{2}\right),
\end{align*}
where it is different from the common interpretation of gradient descent in the way the quadratic approximation is taken by a scaled norm. When $\bL_{t}\approx\bL_{\star}$ and $\bR_{t}\approx\bR_{\star}$ are approaching the ground truth, the additional scaling factors can be approximated by $\bL_{t}^{\top}\bL_{t}\approx\bSigma_{\star}$ and $\bR_{t}^{\top}\bR_{t}\approx\bSigma_{\star}$, leading to the following error metric
\begin{align}
\dist^{2}(\bF,\bF_{\star})\coloneqq\inf_{\bQ\in\GL(r)}\;\left\Vert (\bL\bQ-\bL_{\star})\bSigma_{\star}^{1/2}\right\Vert _{\fro}^{2}+\left\Vert (\bR\bQ^{-\top}-\bR_{\star})\bSigma_{\star}^{1/2}\right\Vert _{\fro}^{2}.\label{eq:dist}
\end{align}
The design and analysis of this new distance metric are of crucial importance in obtaining the improved rate of \texttt{ScaledGD}.
In comparison, the previously studied distance metrics (proposed mainly for GD) either do not include the diagonal scaling \cite{ma2021beyond,tu2015low}, or only consider the ambiguity class up to orthonormal transforms \cite{tu2015low}, which fail to unveil the benefit of \texttt{ScaledGD}.

%% file: tensor.tex
\section{ScaledGD for Low-rank Tensor Estimation}
\label{sec:tensor}


 This section is devoted to introducing \texttt{ScaledGD} and establishing its statistical and computational guarantees for various low-rank tensor estimation problems; the majority of the results are based on \cite{tong2022scaling,dong2023fast}.

\subsection{Assumptions}
\label{sec:models}

Suppose the ground truth tensor $\bcX_{\star} = [ \bcX_{\star}(i_1,i_2,i_3)]\in \RR^{n_1\times n_2\times n_3}$ admits the following Tucker decomposition
\begin{align}\label{eq:Tucker_truth}
\bcX_{\star} (i_1,i_2,i_3) = \sum_{j_1=1}^{r_1}\sum_{j_2=1}^{r_2}\sum_{j_3=1}^{r_3} \bU_{\star} (i_1,j_1) \bV_{\star}(i_2,j_2) \bW_{\star} (i_3,j_3) \bcG_{\star}(j_1,j_2,j_3)
\end{align}
for $1\le i_k \le n_k$, or more compactly,
\begin{align}\label{eq:Tucker_truth_compact}
\bcX_{\star} &=  (\bU_{\star},\bV_{\star},\bW_{\star})\bcdot\bcG_{\star},
\end{align} 
where $\bcG_{\star}=[\bcG_{\star}(j_1,j_2,j_3)]\in \RR^{r_1\times r_2\times r_3}$ is the core tensor of multilinear rank $\br=(r_1,r_2,r_3)$, and $\bU_{\star}=[ \bU_{\star} (i_1,j_1)] \in\RR^{n_1\times r_1}$, $\bV_{\star}=[ \bV_{\star}(i_2,j_2)] \in\RR^{n_2\times r_2}$, $\bW_{\star}=[ \bW_{\star} (i_3,j_3)] \in\RR^{n_3\times r_3}$ are the factor matrices of each mode. Letting $\cM_k(\bcX_{\star})$ be the mode-$k$ matricization of $\bcX_{\star}$, we have
\begin{subequations}\label{eq:matricization}
\begin{align}
\cM_1(\bcX_{\star}) &= \bU_{\star}\cM_1(\bcG_{\star})(\bW_{\star}\otimes\bV_{\star})^{\top}, \\ \cM_2(\bcX_{\star}) &= \bV_{\star}\cM_2(\bcG_{\star})(\bW_{\star}\otimes\bU_{\star})^{\top}, \\ \cM_3(\bcX_{\star}) &= \bW_{\star}\cM_3(\bcG_{\star})(\bV_{\star}\otimes\bU_{\star})^{\top}.
\end{align}
\end{subequations}
It is straightforward to see that the Tucker decomposition is not uniquely specified: for any invertible matrices $\bQ_k \in\RR^{r_k\times r_k}$, $k=1,2,3$, one has 
\begin{align*}
(\bU_{\star},\bV_{\star},\bW_{\star})\bcdot\bcG_{\star} =(\bU_{\star}\bQ_1,\bV_{\star}\bQ_2,\bW_{\star}\bQ_3)\bcdot ((\bQ_1^{-1},\bQ_2^{-1},\bQ_3^{-1}) \bcdot \bcG_{\star}).
\end{align*}
We shall fix the ground truth factors such that $\bU_{\star}$, $\bV_{\star}$ and $\bW_{\star}$ are orthonormal matrices consisting of left singular vectors in each mode. Furthermore, the core tensor $\bcS_{\star}$ is related to the singular values in each mode as
\begin{align}\label{eq:ground_truth_condition}
\cM_{k}(\bcG_{\star})\cM_{k}(\bcG_{\star})^{\top} = \bSigma_{\star,k}^2, \qquad k=1,2,3,
\end{align}
where $\bSigma_{\star,k} \coloneqq \diag[\sigma_{1}(\cM_{k}(\bcX_{\star})),\dots,\sigma_{r_k}(\cM_{k}(\bcX_{\star}))]$ is a diagonal matrix where the diagonal elements are composed of the nonzero singular values of $\cM_{k}(\bcX_{\star})$  and $r_k = \rank(\cM_k(\bcX_{\star}))$ for $k=1,2,3$.

\runinhead{Key parameters.}
Of particular interest is a sort of condition number of $\bcX_{\star}$,  which plays an important role in governing the computational efficiency of first-order algorithms. 

\begin{definition}[Tensor condition number]\label{def:kappa} The condition number of $\bcX_{\star}$ is defined as
\begin{align} \label{eq:kappa}
\kappa \coloneqq \frac{\sigma_{\max}(\bcX_{\star})}{\sigma_{\min}(\bcX_{\star})} = \frac{\max_{k=1,2,3}  \sigma_{1}(\cM_k(\bcX_{\star}))}{\min_{k=1,2,3} \sigma_{r_k}(\cM_k(\bcX_{\star}))}.
\end{align}
\end{definition}

Another parameter is the incoherence parameter, which plays an important role in governing the well-posedness of low-rank tensor RPCA and completion.
\begin{definition}[Tensor incoherence]\label{def:mu} The incoherence parameter of $\bcX_{\star}$ is defined as 
\begin{align}
\mu \coloneqq \max\left\{\frac{n_1}{r_1}\|\bU_{\star}\|_{2,\infty}^2, \; \frac{n_2}{r_2}\|\bV_{\star}\|_{2,\infty}^2, \; \frac{n_3}{r_3}\|\bW_{\star}\|_{2,\infty}^2\right\}.\label{eq:mu}
\end{align}
\end{definition}
Roughly speaking, a small incoherence parameter ensures that the energy of the tensor is evenly distributed across its entries, so that a small random subset of its elements still reveals substantial information about the latent structure of the entire tensor.

\subsection{Tensor sensing}

\runinhead{Observation model.} We first consider tensor sensing --- also known as tensor regression --- with Gaussian design. Assume that we have access to a set of observations given as 
\begin{align}
y_i = \langle\bcA_{i},\bcX_{\star}\rangle, \quad  i=1,\dots,m, \quad \mbox{ or concisely, } \qquad \by = \cA(\bcX_{\star}),
\end{align}
where $\bcA_i \in \RR^{n_1\times n_2 \times n_3}$ is the $i$-th measurement tensor composed of i.i.d.~Gaussian entries drawn from $\cN(0,1/m)$, and $\cA(\bcX) = \{ \langle\bcA_{i},\bcX\rangle \}_{i=1}^m$ is a linear map from $\RR^{n_1\times n_2 \times n_3}$ to $\RR^m$, whose adjoint operator is given by $\cA^*(\by) = \sum_{i=1}^m y_i \bcA_i$. The goal is to recover $\bcX_{\star}$ from $\by$, by leveraging the low-rank structure of $\bcX_{\star}$. 

\runinhead{Algorithm development.} It is natural to minimize the following loss function
\begin{align}\label{eq:loss_TS}
\minimize_{\bF=(\bU,\bV,\bW,\bcG)}\; \cL(\bF)\coloneqq\frac{1}{2}\left\|\cA((\bU,\bV,\bW)\bcdot\bcG)-\by\right\|_{2}^2.
\end{align}
The proposed \texttt{ScaledGD}  algorithm to minimize \eqref{eq:loss_TS} is described in Algorithm~\ref{alg:TR}, where the algorithm is initialized by applying HOSVD to $\cA^*(\by)$, followed by scaled gradient updates given in \eqref{eq:iterates_TR}. 

\begin{algorithm}[h]
\caption{\texttt{ScaledGD} for low-rank tensor sensing}\label{alg:TR} 
\begin{algorithmic} 
\STATE \textbf{Input parameters:} step size $\eta$, multilinear rank $\br = (r_1,r_2,r_3)$. 
\STATE \textbf{Spectral initialization:} Let $\big( \bU_{0},\bV_{0},\bW_{0}, \bcG_{0} \big)$ be the factors in the top-$\br$ HOSVD of $\cA^{*}(\by)$ (cf. \eqref{eq:HOSVD}).
\STATE \textbf{Scaled gradient updates:} for $t=0,1,2,\dots,T-1$  
\begin{align}
\begin{split} 
\bU_{t+1} &= \bU_{t}-\eta \cM_{1}\left(\cA^{*}(\cA((\bU_{t},\bV_{t},\bW_{t})\bcdot\bcG_{t})-\by)\right)\breve{\bU}_t^{\top} \big(\breve{\bU}_t^{\top} \breve{\bU}_t \big)^{-1}, \\
\bV_{t+1} &= \bV_{t}-\eta \cM_{2}\left(\cA^{*}(\cA((\bU_{t},\bV_{t},\bW_{t})\bcdot\bcG_{t})-\by)\right)\breve{\bV}_t^{\top}\big(\breve{\bV}_t^{\top} \breve{\bV}_t \big)^{-1}, \\
\bW_{t+1} &= \bW_{t} - \eta \cM_{3}\left(\cA^{*}(\cA((\bU_{t},\bV_{t},\bW_{t})\bcdot\bcG_{t})-\by)\right)\breve{\bW}_t^{\top}\big(\breve{\bW}_t^{\top} \breve{\bW}_t \big)^{-1}, \\
\bcG_{t+1} &= \bcG_{t} - \eta\left((\bU_{t}^{\top}\bU_{t})^{-1}\bU_{t}^{\top},(\bV_{t}^{\top}\bV_{t})^{-1}\bV_{t}^{\top},(\bW_{t}^{\top}\bW_{t})^{-1}\bW_{t}^{\top}\right) \\& \qquad \qquad\qquad\qquad \bcdot \cA^{*}(\cA((\bU_{t},\bV_{t},\bW_{t})\bcdot\bcG_{t})-\by),
\end{split}\label{eq:iterates_TR}
\end{align}
where $\breve{\bU}_t$, $\breve{\bV}_t$, and $\breve{\bW}_t$ are defined in \eqref{eq:breve_uvw}.
\end{algorithmic} 
\end{algorithm}

\runinhead{Theoretical guarantee.} Encouragingly, we can guarantee that \texttt{ScaledGD} provably recovers the ground truth tensor as long as the sample size is sufficiently large, which is given in the following theorem.

\begin{theorem} \label{thm:TR} Let $n = \max_{k=1,2,3} n_k$ and $r =\max_{k=1,2,3}r_k$. With Gaussian design, suppose that $m$ satisfies 
\begin{align*}
m \gtrsim \epsilon_{0}^{-1}\sqrt{n_1n_2n_3}r^{3/2}\kappa^2 + \epsilon_{0}^{-2}(nr^2\kappa^4\log n + r^{4}\kappa^{2})
\end{align*}
for some small constant $\epsilon_0>0$. If the step size obeys $0 < \eta \le 2/5$, then with probability at least $1-c_{1}n^{-c_{2}}$ for universal constants $c_1,c_2>0$, for all $t\ge 0$, the iterates of Algorithm~\ref{alg:TR} satisfy
\begin{align*}
\left\|(\bU_{t},\bV_{t},\bW_{t})\bcdot\bcG_{t}-\bcX_{\star}\right\|_{\fro} & \le 3\epsilon_0 (1-0.6\eta)^{t}\sigma_{\min}(\bcX_{\star}).
\end{align*}
\end{theorem}
Theorem~\ref{thm:TR} ensures that the reconstruction error $\left\|(\bU_{t},\bV_{t},\bW_{t})\bcdot\bcG_{t}-\bcX_{\star}\right\|_{\fro} $ contracts linearly at a constant rate independent of the condition number of $\bcX_{\star}$; to find an $\varepsilon$-accurate estimate, i.e.~$ \|(\bU_{t},\bV_{t},\bW_{t})\bcdot\bcG_{t}-\bcX_{\star} \|_{\fro} \le \varepsilon \sigma_{\min}(\bcX_{\star})$, \texttt{ScaledGD} needs at most $O(\log (1/\varepsilon))$ iterations, as long as the sample complexity satisfies 
\begin{align*}
m \gtrsim n^{3/2}r^{3/2}\kappa^2,
\end{align*}
where again we keep only terms with dominating orders of $n$.
Compared with regularized GD \cite{han2020optimal}, \texttt{ScaledGD} achieves a low computation complexity with robustness to ill-conditioning, improving its iteration complexity by a factor of $\kappa^2$, and does not require any explicit regularization.

\subsection{Tensor robust principal component analysis}

\runinhead{Observation model.} Suppose that we collect a set of corrupted observations of $\bcX_{\star}$ as
\begin{equation}\label{eq:obs_model}
\bcY = \bcX_{\star} + \bcS_{\star},
\end{equation}
where $\bcS_{\star}$ is the corruption tensor. The problem of tensor RPCA seeks to separate $\bcX_{\star}$ and $\bcS_{\star}$ from their sum $\bcY$ as efficiently and accurately as possible. Similar to the matrix case, we consider a deterministic sparsity model following the matrix case~\cite{chandrasekaran2011siam,netrapalli2014non,yi2016fast}, where $\bcS_{\star}$ contains at most a small fraction of nonzero entries per fiber. Formally, the corruption tensor $\bcS_{\star}$ is said to be $\alpha$-fraction sparse, i.e., $\bcS_{\star}\in\bcS_{\alpha}$, where  
\begin{multline}
\bcS_{\alpha}\coloneqq \Big\{\bcS\in\RR^{n_{1}\times n_{2}\times n_3}:\; \|\bcS_{i_1,i_2,:}\|_{0}\le\alpha n_{3}, \; \|\bcS_{i_1,:,i_3}\|_{0}\le\alpha n_{2}, \;   \\
\|\bcS_{:,i_2,i_3}\|_{0}\le\alpha n_{1},  \mbox{ for all  } 1\leq i_k \leq n_k, \quad k=1,2,3 \Big\}.\label{eq:S_alpha}
\end{multline}




\runinhead{ScaledGD algorithm.} 
It is natural to optimize the following objective function:
\begin{align}\label{eq:loss}
\minimize_{\bF, \bcS}\; \cL(\bF,\bcS) \coloneqq \frac{1}{2}\left\| \big(\bU,\bV,\bW \big)\bcdot \bcG+\bcS-\bcY \right\|_{\fro}^{2},
\end{align}
where $\bF=(\bU,\bV,\bW,\bcG)$ and $\bcS$ are the optimization variables for the tensor factors and the corruption tensor, respectively. 
Our algorithm alternates between corruption removal and factor refinements, as detailed in Algorithm \ref{alg:tensor_RPCA}. To remove the corruption, we use the following soft-shrinkage operator that trims the magnitudes of the entries by the amount of some carefully pre-set threshold.

\begin{definition}[Soft-shrinkage operator]
For an order-$3$ tensor $\bcX$, the soft-shrinkage operator $\Shrink{\zeta}{\cdot}:\, \mathbb{R}^{n_1\times n_2\times n_3} \mapsto \mathbb{R}^{n_1\times n_2\times n_3}$ with threshold $\zeta>0$ is defined as
\begin{align*}
    \big[\Shrink{\zeta}{\bcX}\big]_{i_1, i_2, i_3} := \sgn\big([\bcX]_{i_1, i_2, i_3}\big) \,\cdot\,\max\big(0, \big| [\bcX]_{i_1, i_2, i_3} \big| - \zeta \big) .
\end{align*}
\end{definition}
The soft-shrinkage operator sets entries with magnitudes smaller than $\zeta$ to $0$, while uniformly shrinking the magnitudes of the other entries by $\zeta$. At the beginning of each iteration, the corruption tensor is updated via applying the soft-thresholding operator to the current residual $\bcY - \big(\bU_t, \bV_t, \bW_t \big) \bcdot \bcG_t$ using some properly selected threshold $\zeta_t$, followed by updating the tensor factors $\bF_t$ via scaled gradient descent with respect to $\cL(\bF_t, \bcS_{t+1})$ in \eqref{eq:loss}. To complete the algorithm description, we still need to specify how to initialize the algorithm. This is again achieved by the spectral method, which computes the rank-$\br$ HOSVD of the observation after applying the soft-shrinkage operator


\begin{algorithm}[t]
\caption{\texttt{ScaledGD} for tensor robust principal component analysis}\label{alg:tensor_RPCA} 
\begin{algorithmic} 
\STATE \textbf{Input:} observed tensor $\bcY$, multilinear rank $\br$, learning rate $\eta$, and threshold schedule $\{\zeta_t\}_{t=0}^{T}$. 
\STATE \textbf{Spectral initialization:} Set $\bcS_0 = \Shrink{\zeta_0}{\bcY}$ and $\big(\bU_0, \bV_0, \bW_0, \bcG_0 \big)$ as the factors in the top-$\br$ HOSVD of $\bcY -\bcS_0$  (cf. \eqref{eq:HOSVD}).
\STATE \textbf{{Scaled gradient updates}}: \textbf{for} $t=0,1,2,\dots,T-1$ \textbf{do} 
\begin{subequations}
\begin{align}
\bcS_{t+1} & = \Shrink{\zeta_{t+1}}{\bcY - \big(\bU_t, \bV_t, \bW_t \big) \bcdot \bcG_t},  \\ 
  \bU_{t+1}     &= \bU_t - \eta \left( \bU_t \Breve{\bU}_t^{\top}   + \Matricize{1}{\bcS_{t+1}} -  \Matricize{1}{\bcY}\right) \Breve{\bU}_t \big(\Breve{\bU}_t^{\top} \Breve{\bU}_t \big)^{-1}  , \\
    \bV_{t+1}     &= \bV_t - \eta \left( \bV_t \Breve{\bV}_t^{\top}   + \Matricize{2}{\bcS_{t+1}} -  \Matricize{2}{\bcY}\right) \Breve{\bV}_t \big(\Breve{\bV}_t^{\top} \Breve{\bV}_t \big)^{-1}  , \\
      \bW_{t+1}     &= \bW_t - \eta \left( \bW_t \Breve{\bW}_t^{\top}   + \Matricize{1}{\bcS_{t+1}} -  \Matricize{1}{\bcY}\right) \Breve{\bW}_t \big(\Breve{\bW}_t^{\top} \Breve{\bW}_t \big)^{-1}  , \\
      \bcG_{t+1}     &= \bcG_t - \eta \left( \big(\bU^{\top}_t \bU_t \big)^{-1} \bU_t^{\top}, \big(\bV^{\top}_t \bV_t\big)^{-1} \bV_t^{\top}, \big(\bW^{\top}_t \bW_t\big)^{-1} \bW_t^{\top} \right)  \nonumber \\
    & \qquad \qquad\qquad \bcdot \left( \big(\bU_t, \bV_t, \bW_t \big) \bcdot \bcG_t + \bcS_{t+1} - \bcY \right) ,
\end{align}
\end{subequations}
where $\breve{\bU}_t$, $\breve{\bV}_t$, and $\breve{\bW}_t$ are defined in \eqref{eq:breve_uvw}.
\end{algorithmic} 
\end{algorithm}



\runinhead{Theoretical guarantee.}
Fortunately, the \texttt{ScaledGD} algorithm provably recovers the ground truth tensor---as long as the fraction of corruptions is not too large---with proper choices of the tuning parameters, as captured in following theorem.
\begin{theorem} \label{main}
Let $\bcY = \bcX_\star + \bcS_\star \in \RR^{n_1 \times n_2 \times n_3}$, where $\bcX_\star$ is $\mu$-incoherent with multilinear rank $\br = (r_1, r_2, r_3)$, and $\bcS_\star$ is $\alpha$-sparse. Suppose that the threshold values $\{\zeta_k\}_{k=0}^\infty$ obey that $  \norm{\bcX_\star}_\infty \leq \zeta_0  \leq 2     \norm{\bcX_\star}_\infty$ and  $\zeta_{t+1} = \rho \zeta_{t}$, $t\geq 1$, for some properly tuned $\zeta_1: =  8 \sqrt{\frac{\mu^3 r_1 r_2 r_3}{n_1 n_2 n_3}}   \sigma_{\min}(\bcX_\star)
$ and $\frac{1}{7} \leq \eta \leq \frac{1}{4}$, where $\rho = 1-0.45\eta$. Then,  the iterates $\bcX_t =  \big(\bU_t, \bV_t, \bW_t \big) \bcdot \bcG_t$ satisfy
\begin{subequations} \label{eq:thm_claims}
\begin{align}
    \norm{ \bcX_t - \bcX_\star}_{\fro} &\leq 0.03 \rho^t  \sigma_{\min}(\bcX_\star) , \label{eq:fro_contraction}  \\ 
 \norm{ \bcX_t - \bcX_\star}_{\infty}  & \leq 8 \rho^t \sqrt{\frac{\mu^3 r_1 r_2 r_3}{n_1 n_2 n_3}}   \sigma_{\min}(\bcX_\star) \label{eq:entry_contraction} 
\end{align}
\end{subequations}
for all $t\geq 0$, as long as  the level of corruptions obeys
 $\alpha \leq \frac{c_0}{\mu^2 r_1 r_2 r_3 \kappa}$
 for some sufficiently small $c_0>0$.
\end{theorem}

Theorem~\ref{main} implies that upon appropriate choices of the parameters, if the level of corruptions $\alpha$ is small enough, i.e. not exceeding the order of $\frac{1}{\mu^2 r_1 r_2 r_3 \kappa}$, we can ensure that \texttt{ScaledGD} converges at a linear rate independent of the condition number and exactly recovers the ground truth tensor $\bcX_\star$ even when the gross corruptions are arbitrary and adversarial.  Furthermore, when $\mu=O(1)$ and $r =O(1)$, the entrywise error bound \eqref{eq:entry_contraction}---which is smaller than the Frobenius error \eqref{eq:fro_contraction} by a factor of $\sqrt{\frac{1}{n_1n_2n_3}}$---suggests the errors are distributed in an even manner across the entries for incoherent and low-rank tensors. 


\subsection{Tensor completion}

\runinhead{Observation model.} Assume that we have observed a subset of entries in $\bcX_{\star}$, given as 
$\bcY= \cP_{\Omega}(\bcX_{\star})$, where $\cP_{\Omega}:\RR^{n_{1}\times n_{2}\times n_3}\mapsto\RR^{n_{1}\times n_{2}\times n_3}$ is a projection such that
\begin{align}
[\cP_{\Omega}(\bcX_{\star})] (i_1,i_2,i_3) =\begin{cases} \bcX_{\star} (i_1,i_2,i_3), & \mbox{if }(i_1,i_2,i_3)\in\Omega, \\
0, & \mbox{otherwise}.\end{cases}
\end{align}
Here, $\Omega$ is generated according to the Bernoulli observation model in the sense that 
\begin{align} \label{eq:bernoulli_model}
(i_1,i_2,i_3) \in \Omega ~~\mbox{ independently with probability}~p\in (0,1].
\end{align}
The goal  is to recover the tensor $\bcX_{\star}$ from its partial observation $\cP_{\Omega}(\bcX_{\star})$, which can be achieved by minimizing the loss function
\begin{align}\label{eq:loss_TC}
\min_{\bF=(\bU,\bV,\bW,\bcS)}\; \cL(\bF)\coloneqq\frac{1}{2p}\left\|\cP_{\Omega}\big((\bU,\bV,\bW)\bcdot\bcS \big) - \bcY \right\|_{\fro}^2. 
\end{align}

 \runinhead{Algorithm development.}
To guarantee faithful recovery from partial observations, the underlying low-rank tensor $\bcX_{\star}$ needs to be incoherent (cf.~Definition~\ref{def:mu}) to avoid ill-posedness. One typical strategy, as employed in the matrix setting, to ensure the incoherence condition is to trim the rows of the factors after the scaled gradient update. We introduce the scaled projection as follows, 
 \begin{align}\label{eq:scaled_proj}
(\bU, \bV, \bW, \bcS) = \cP_{B}(\bU_{+},{\bV}_{+},{\bW}_{+}, {\bcS}_{+}),\
\end{align}
where $B>0$ is the projection radius, and
\begin{align*}
\bU(i_1,:) & = \left(1 \wedge \frac{B}{\sqrt{n_1}  \| {\bU}_{+}(i_1,:) \breve{\bU}_{+}^{\top}  \|_2}\right)  \bU_+(i_1,:), \qquad 1\le i_1\le n_1; \\
\bV(i_2,:) & = \left(1 \wedge \frac{B}{\sqrt{n_2}\| {\bV}_{+}(i_2,:) \breve{\bV}_{+}^{\top}\|_2}\right) \bV_{+}(i_2,:), \qquad 1\le i_2\le n_2; \\
\bW(i_3,:)  &= \left(1 \wedge \frac{B}{\sqrt{n_3}\| \bW_{+}(i_3,:) \breve{\bW}_{+}^{\top}\|_2}\right) {\bW}_{+}(i_3,:),  \qquad 1\le i_3\le n_3; \\
\bcS & = \bcS_{+}.
\end{align*}
Here, we recall $\breve{\bU}_{+}$, $\breve{\bV}_{+}$, $\breve{\bW}_{+}$ are analogously defined in \eqref{eq:breve_uvw} using $(\bU_{+}, \bV_{+},\bW_{+},\bcS_{+})$.
As can be seen, each row of ${\bU}_{+}$ (resp.~${\bV}_{+}$ and ${\bW}_{+}$) is scaled by a scalar based on the row $\ell_2$ norms of ${\bU}_{+}\breve{\bU}_{+}^{\top}$ (resp.~${\bV}_{+}\breve{\bV}_{+}^{\top}$ and ${\bW}_{+}\breve{\bW}_{+}^{\top}$), which is the mode-1 (resp.~mode-2 and mode-3) matricization of the tensor $(\bU_{+}, \bV_{+}, \bW_{+} )\bcdot\bcS_{+}$. It is a straightforward observation that the projection can be computed efficiently.

With the scaled projection $\cP_B(\cdot)$ defined in hand, we are in a position to describe the details of the proposed \texttt{ScaledGD} algorithm, summarized in Algorithm~\ref{alg:TC}. It consists of two stages: spectral initialization followed by iterative refinements using the scaled projected gradient updates in 
\eqref{eq:iterates_TC}. For the spectral initialization, we take advantage of the subspace estimators proposed in \cite{cai2019subspace,xia2021statistically} for highly unbalanced matrices. Specifically, we estimate the subspace spanned by $\bU_{\star}$ by that spanned by top-$r_1$ eigenvectors $\bU_{+}$ of the diagonally-deleted Gram matrix of $p^{-1}\cM_{1}(\bcY)$, denoted as
\begin{align*}
\Poffdiag(p^{-2}\cM_{1}(\bcY)\cM_{1}(\bcY)^{\top}),
\end{align*}
and the other two factors $\bV_{+}$ and $\bW_{+}$ are estimated similarly. The core tensor is then estimated as
\begin{align*}
\bcS_{+} = p^{-1} (\bU_{+}^{\top},\bV_{+}^{\top},\bW_{+}^{\top})\bcdot \bcY.
\end{align*}
To ensure the initialization is incoherent, we pass it through the scaled projection operator to obtain the final initial estimate:
\begin{align*}
(\bU_{0},\bV_{0},\bW_{0},\bcS_{0})  = \cP_{B} \big(\bU_{+},\bV_{+},\bW_{+},\bcS_{+}\big).
\end{align*}

\begin{algorithm}[t]
\caption{\texttt{ScaledGD} for low-rank tensor completion}\label{alg:TC} 
\begin{algorithmic} 
\STATE \textbf{Input parameters:} step size $\eta$, multilinear rank $\br = (r_1,r_2,r_3)$, probability of observation $p$, projection radius $B$.
\STATE \textbf{Spectral initialization:} Let $\bU_{+}$ be the top-$r_1$ eigenvectors of
$\Poffdiag(p^{-2}\cM_{1}(\bcY)\cM_{1}(\bcY)^{\top})$,
and similarly for $\bV_{+},\bW_{+}$, and $\bcS_{+} = p^{-1} (\bU_{+}^{\top},\bV_{+}^{\top},\bW_{+}^{\top})\bcdot \bcY$. 
Set $(\bU_{0},\bV_{0},\bW_{0},\bcS_{0})  = \cP_{B} \big(\bU_{+},\bV_{+},\bW_{+},\bcS_{+}\big)$.
\STATE \textbf{Scaled gradient updates:} for $t=0,1,2,\dots,T-1$, \textbf{do}  
\begin{align}
\begin{split} 
{\bU}_{t+} &= \bU_{t}- \frac{\eta}{p}\cM_{1}\left(\cP_{\Omega}\big((\bU_{t},\bV_{t},\bW_{t})\bcdot\bcS_{t} \big) - \bcY \right)\breve{\bU}_t \big(\breve{\bU}_t^{\top} \breve{\bU}_t \big)^{-1}, \\
{\bV}_{t+} &= \bV_{t}- \frac{\eta}{p}\cM_{2}\left(\cP_{\Omega}\big((\bU_{t},\bV_{t},\bW_{t})\bcdot\bcS_{t} \big) - \bcY \right)\breve{\bV}_t\big(\breve{\bV}_t^{\top} \breve{\bV}_t \big)^{-1}, \\
{\bW}_{t+} &= \bW_{t} -  \frac{\eta}{p}\cM_{3}\left(\cP_{\Omega}\big((\bU_{t},\bV_{t},\bW_{t})\bcdot\bcS_{t} \big) - \bcY \right)\breve{\bW}_t\big(\breve{\bW}_t^{\top} \breve{\bW}_t \big)^{-1}, \\
{\bcS}_{t+} &= \bcS_{t} - \frac{\eta}{p}\left((\bU_{t}^{\top}\bU_{t})^{-1}\bU_{t}^{\top},(\bV_{t}^{\top}\bV_{t})^{-1}\bV_{t}^{\top},(\bW_{t}^{\top}\bW_{t})^{-1}\bW_{t}^{\top}\right) \\
& \qquad \qquad\qquad\qquad \bcdot  \left(\cP_{\Omega}\big((\bU_{t},\bV_{t},\bW_{t})\bcdot\bcS_{t} \big) - \bcY \right),
\end{split}\label{eq:iterates_TC}
\end{align}
where $\breve{\bU}_t$, $\breve{\bV}_t$, and $\breve{\bW}_t$ are defined in \eqref{eq:breve_uvw}. Set
$$(\bU_{t+1},\bV_{t+1},\bW_{t+1},\bcS_{t+1})  = \cP_{B}({\bU}_{t+}, {\bV}_{t+}, {\bW}_{t+}, {\bcS}_{t+}) .$$
\end{algorithmic} 
\end{algorithm}

\runinhead{Theoretical guarantee.} The following theorem establishes the performance guarantee of \texttt{ScaledGD} for tensor completion, as soon as the sample size is sufficiently large.

\begin{theorem}\label{thm:TC} Let $n = \max_{k=1,2,3} n_k$ and $r =\max_{k=1,2,3}r_k$. Suppose that $\bcX_{\star}$ is $\mu$-incoherent, $n_k\gtrsim \epsilon_{0}^{-1}\mu r_k^{3/2}\kappa^2$ for $k=1,2,3$, and that $p$ satisfies 
\begin{align*}
pn_1n_2n_3 \gtrsim \epsilon_{0}^{-1}\sqrt{n_1n_2n_3}\mu^{3/2}r^{5/2}\kappa^{3} \log^{3} n + \epsilon_{0}^{-2}n\mu^{3} r^{4}\kappa^{6}\log^{5} n
\end{align*}
for some small constant $\epsilon_0>0$. Set the projection radius as $B=C_{B}\sqrt{\mu r}\sigma_{\max}(\bcX_{\star})$ for some constant $C_{B}\ge (1+\epsilon_{0})^{3}$. If the step size obeys $0<\eta\le2/5$, then with probability at least $1-c_{1}n^{-c_{2}}$ for universal constants $c_1,c_2>0$, for all $t\ge0$, the iterates of Algorithm~\ref{alg:TC} satisfy
\begin{align*}
\left\|(\bU_{t},\bV_{t},\bW_{t})\bcdot\bcS_{t}-\bcX_{\star}\right\|_{\fro}\le3\epsilon_{0}(1-0.6\eta)^{t}\sigma_{\min}(\bcX_{\star}).
\end{align*} 
\end{theorem}

Theorem~\ref{thm:TC} ensures that to find an $\varepsilon$-accurate estimate, i.e.~$ \|(\bU_{t},\bV_{t},\bW_{t})\bcdot\bcS_{t}-\bcX_{\star} \|_{\fro} \le \varepsilon \sigma_{\min}(\bcX_{\star})$, \texttt{ScaledGD} takes at most $O(\log (1/\varepsilon))$ iterations, which is independent of the condition number of $\bcX_{\star}$, as long as the sample complexity is large enough. Assuming that $\mu= O(1)$ and $r \vee \kappa\ll n^{\delta}$ for some small constant $\delta$ to keep only terms with dominating orders of $n$, the sample complexity simplifies to
\begin{align*}
pn_1n_2n_3 \gtrsim n^{3/2}r^{5/2}\kappa^{3}\log^3 n,
\end{align*}
which is near-optimal in view of the conjecture that no polynomial-time algorithm will be successful if the sample complexity is less than the order of $n^{3/2}$ for tensor completion \cite{barak2016noisy}. Compared with existing algorithms collected in Table~\ref{tab:ScaledGD-tensor-completion}, \texttt{ScaledGD} is the {\em first} algorithm that simultaneously achieves a near-optimal sample complexity and a near-linear run time complexity in a provable manner. 

\begin{table}[t]
\centering %
\begin{tabular}{c||c|c|c}
\hline 
Algorithms & Sample complexity & Iteration complexity & Parameter space   \tabularnewline
\hline  \hline 
Unfolding + nuclear norm min.  & \multirow{2}{*}{$n^2 r \log^2 n$} & \multirow{2}{*}{polynomial} & \multirow{2}{*}{tensor}  \tabularnewline 
\cite{huang2015provable} &  &  &  \tabularnewline \hline 
Tensor nuclear norm min.  & \multirow{2}{*}{$ n^{3/2}r^{1/2} \log^{3/2} n$} & \multirow{2}{*}{NP-hard} & \multirow{2}{*}{tensor}  \tabularnewline
\cite{yuan2016tensor} &  &  &  \tabularnewline \hline 
Grassmannian GD  & \multirow{2}{*}{$n^{3/2}r^{7/2}\kappa^4 \log^{7/2}n$} & \multirow{2}{*}{N/A} & \multirow{2}{*}{factor}  \tabularnewline
\cite{xia2019polynomial} &  &  &  \tabularnewline \hline 
\multirow{2}{*}{\texttt{ScaledGD}}  & \multirow{2}{*}{$n^{3/2}r^{5/2}\kappa^{3}\log^3 n$ } & \multirow{2}{*}{$\log\frac{1}{\varepsilon}$} & \multirow{2}{*}{factor}  \tabularnewline
  &  &  &  \tabularnewline
\hline
\end{tabular}\vspace{0.04in}
\caption{Comparisons of ScaledGD with existing algorithms for tensor completion when the tensor is incoherent and low-rank under the Tucker decomposition. Here, we say that the output $\bcX$ of an algorithm reaches $\varepsilon$-accuracy, if it satisfies $\|\bcX-\bcX_{\star}\|_{\fro}\le\varepsilon\sigma_{\min}(\bcX_{\star})$. Here, $\kappa$ and $\sigma_{\min}(\bcX_{\star})$ are the condition number and the minimum singular value of $\bcX_{\star}$ (defined in Section~\ref{sec:models}).  For simplicity, we let $n = \max_{k=1,2,3} n_k$ and $r =\max_{k=1,2,3}r_k$, and assume $r\vee \kappa\ll n^{\delta}$ for some small constant $\delta$ to keep only terms with dominating orders of $n$. 
\label{tab:ScaledGD-tensor-completion}  }
\end{table}

%% file: overparam.tex
\section{Preconditioning Meets Overparameterization}
\label{sec:overparam}
In this section we treat a more complicated scenario, 
where the correct rank $r$ of the ground truth $\bX_\star$ is not known \emph{a priori}. In this case, a practical solution is to \emph{overparameterize}, i.e., to choose some $r'>r$, and proceed as if $r'$ is the correct rank. 
It turns out that \texttt{ScaledGD} needs some simple modification to work robustly in this setting. The modified algorithm, which we call \texttt{ScaledGD($\lambda$)}, will be introduced in the rest of this section, along with theoretical analysis on its global convergence.

\runinhead{Motivation.}
We begin with inspecting the behavior of \texttt{ScaledGD} in the overparameterized setting. Assume we already find some $\bL_t\in\RR^{n_1\times r'}, \bR_t\in\RR^{n_2\times r'}$ that are close to the ground truth: the first $r$ columns of $\bL_t$ 
are close to $\bL_\star$, while the rest $r'-r$ columns are close to zero; \emph{mutatis mutandis} for $\bR_t$.

Recall that in the update equation~\eqref{eq:scaledGD} of \texttt{ScaledGD}:
\begin{align*}
\begin{split} \bL_{t+1} & =\bL_{t}-\eta\nabla_{\bL}\cL(\bL_{t},\bR_{t})(\bR_{t}^{\top}\bR_{t})^{-1},\\
    \bR_{t+1} & =\bR_{t}-\eta\nabla_{\bR}\cL(\bL_{t},\bR_{t})(\bL_{t}^{\top}\bL_{t})^{-1},
\end{split}
\end{align*}
the preconditioners are chosen to be the inverse of the $r'\times r'$ matrices $\bL_t^\top\bL_t$, $\bR_t^\top\bR_t$. However, since the last $r'-r$ columns for $\bL_t$ (respectively, $\bR_t$) are close to zero, it is clear that $\bL_t^\top\bL_t$ (respectively, $\bR_t^\top\bR_t$) are approximately of rank at most $r$. When $r'>r$, this means $\bL_t^\top\bL_t$ and $\bR_t^\top\bR_t$ are close to being degenerate since their approximate ranks (no larger than $r$) are smaller than their dimensions $r'$, thus taking inverse of them are numerically unstable.

\runinhead{\texttt{ScaledGD($\lambda$)}: \texttt{ScaledGD} with overparameterization.}
One of the simplest remedies to such instability is to \emph{regularize} the preconditioner. Before taking inverse of $\bL_t^\top\bL_t$ and $\bR_t^\top\bR_t$, we add a regularizer $\lambda\bI$ to avoid degeneracy, where $\lambda>0$ is a regularization parameter; the preconditioner thus becomes $(\bL_t^\top\bL_t+\lambda\bI)^{-1}$ and $(\bR_t^\top\bR_t+\lambda\bI)^{-1}$. 
The new update rule is specified by
\begin{align}
\begin{split} 
    \bL_{t+1} & =\bL_{t}-\eta\nabla_{\bL}\cL(\bL_{t},\bR_{t})(\bR_{t}^{\top}\bR_{t} + \lambda\bI)^{-1},\\
    \bR_{t+1} & =\bR_{t}-\eta\nabla_{\bR}\cL(\bL_{t},\bR_{t})(\bL_{t}^{\top}\bL_{t} + \lambda\bI)^{-1}.
    \label{eq:overpar-scaledGD}
\end{split}
\end{align}

This regularized version of \texttt{ScaledGD} is called \texttt{ScaledGD($\lambda$)}.
It turns out that the simple regularization trick works out well: \texttt{ScaledGD($\lambda$)} not only (almost) inherits the $\kappa$-free convergence rate, but also has the advantage of being robust to overparameterization and enjoying provable global convergence to any prescribed accuracy level, when initialized by a small random initialization. This is established in \cite{xu2023power} formally for the matrix sensing setting studied in Section~\ref{subsec:scaledGD_MC}, where  the two factors $\bL_\star$, $\bR_\star$ in $\bX_\star=\bL_\star\bR_\star^\top$ are equal, i.e. $\bX_\star=\bL_\star\bL_\star^\top$. Under such situation, \texttt{ScaledGD($\lambda$)} instantiates to 
\begin{equation}
 \bL_{t+1}=\bL_{t}-\eta\cA^{*}(\cA(\bL_{t}\bL_{t}^{\top})-\by)\bL_{t}(\bL_{t}^{\top}\bL_{t} + \lambda\bI )^{-1}.    \label{eq:overpar-scaledGD-sym}
\end{equation}
Similar to \cite{li2018algorithmic,stoger2021small}, 
we set the initialization $\bL_0$ to be a small random matrix, i.e.,
\begin{equation}
\bL_0=\alpha \bG,
\end{equation}
where $\bG \in \mathbb{R}^{n\times r'}$ is some matrix considered to be normalized 
and $\alpha>0$ controls the magnitude of the initialization.
To simplify exposition, we take $\bG$ to be a standard random Gaussian matrix, 
that is, $\bG$ is a random matrix with i.i.d. entries distributed as $\mathcal{N}(0,1/n)$.


\runinhead{Theoretical guarantee.} We have the performance guarantee of \texttt{ScaledGD($\lambda$)} under the standard RIP assumption as follows.

\begin{theorem}\label{thm:main} 
Suppose that $\mathcal{A}(\cdot)$ satisfies the rank-$(r+1)$ RIP 
with $\delta_{r+1} \eqqcolon \delta$. Furthermore, there exist a sufficiently small constant $c_\delta>0$ and a sufficiently large constant $C_\delta > 0$ such that
\begin{equation}     \label{eqn:delta-cond}
    \delta \le c_\delta r^{-1/2}\kappa^{- C_\delta }.
\end{equation}    
Assume there exist some universal constants 
$c_\eta, c_\lambda ,  C_\alpha>0$ 
such that $(\eta, \lambda, \alpha)$ in \texttt{ScaledGD($\lambda$)}
satisfy the following conditions:
\begin{subequations} \label{eq:param_conditions}
    \begin{align}
\mathsf{(learning~rate)} & \qquad\qquad\qquad  \eta  \le c_\eta, 
    \label{eqn:eta-cond}\\
    \mathsf{(damping~parameter)} & \qquad    \frac{1}{100}c_\lambda\sigma_{\min}^2(\bL_\star)  \le\lambda \le c_\lambda\sigma_{\min}^2(\bL_\star), 
    \label{eqn:lambda-cond}\\
    \mathsf{(initialization~size)} & \qquad     \log\frac{\|\bL_\star\|}{\alpha}  \ge \frac{C_\alpha}{\eta}\log(2\kappa)\cdot\log(2\kappa n). 
    \label{eqn:alpha-cond}
    \end{align}
\end{subequations}
With high probability (with respect to the realization of the random initialization $\bG$),
there exists a universal constant $C_{\min} > 0$ such that
for some $T\le T_{\min} \coloneqq \frac{C_{\min}}{\eta}\log\frac{\|\bL_\star\|}{\alpha}$, the iterates $\bL_t$ of~\eqref{eq:overpar-scaledGD-sym}, obey 
\[
    \|\bL_T \bL_T^\top - \bX_\star\|_{\fro}\le\alpha^{1/3}\|\bL_\star\|^{5/3}.
\]
In particular, for any prescribed accuracy target $\epsilon\in(0,1)$, 
by choosing a sufficiently small $\alpha$ fulfilling both~\eqref{eqn:alpha-cond} 
and $\alpha\le\epsilon^3\|\bL_\star\|$, we have
$\|\bL_T\bL_T^\top - \bX_\star\|_{\fro}\le \epsilon\|\bX_\star\|$.
\end{theorem}

Theorem~\ref{thm:main} shows that by choosing an appropriate $\alpha$, \texttt{ScaledGD($\lambda$)} finds an $\epsilon$-accurate solution, i.e., $\|\bL_T\bL_T^{\top}- \bX_\star \|_{\fro}\le \epsilon\|  \bX_\star \|$, in no more than an order of
$$ \log\kappa \cdot \log (\kappa n) + \log(1/\epsilon)$$ iterations. Roughly speaking, this asserts that \texttt{ScaledGD($\lambda$)} converges at a constant linear rate independent of the condition number $\kappa$ after an initial phase of approximately $O(\log\kappa \cdot \log (\kappa n) )$ iterations. In contrast, overparameterized \texttt{GD} requires $O(\kappa^4 + \kappa^3\log(\kappa n/\epsilon) )$ iterations to converge from a small random initialization to $\epsilon$-accuracy; see~\cite{stoger2021small, li2018algorithmic}. Thus, the convergence of overparameterized \texttt{GD} is much slower than \texttt{ScaledGD($\lambda$)} even for mildly ill-conditioned matrices. Furthermore,
 our sample complexity depends only on the true rank $r$, but not on the overparameterized rank $r'$ --- a crucial feature in order to provide meaningful guarantees when the overparameterized rank $r'$ is uninformative of the true rank, i.e., close to the full dimension $n$. 
 The dependency on $\kappa$ in the sample complexity, on the other end, has been generally unavoidable in nonconvex low-rank estimation \cite{chi2019nonconvex}.

%% file: numerical.tex
\section{Numerical Experiments}
 \label{sec:numerical}
 
\begin{figure}[h]
\centering
\includegraphics[width=0.5\textwidth]{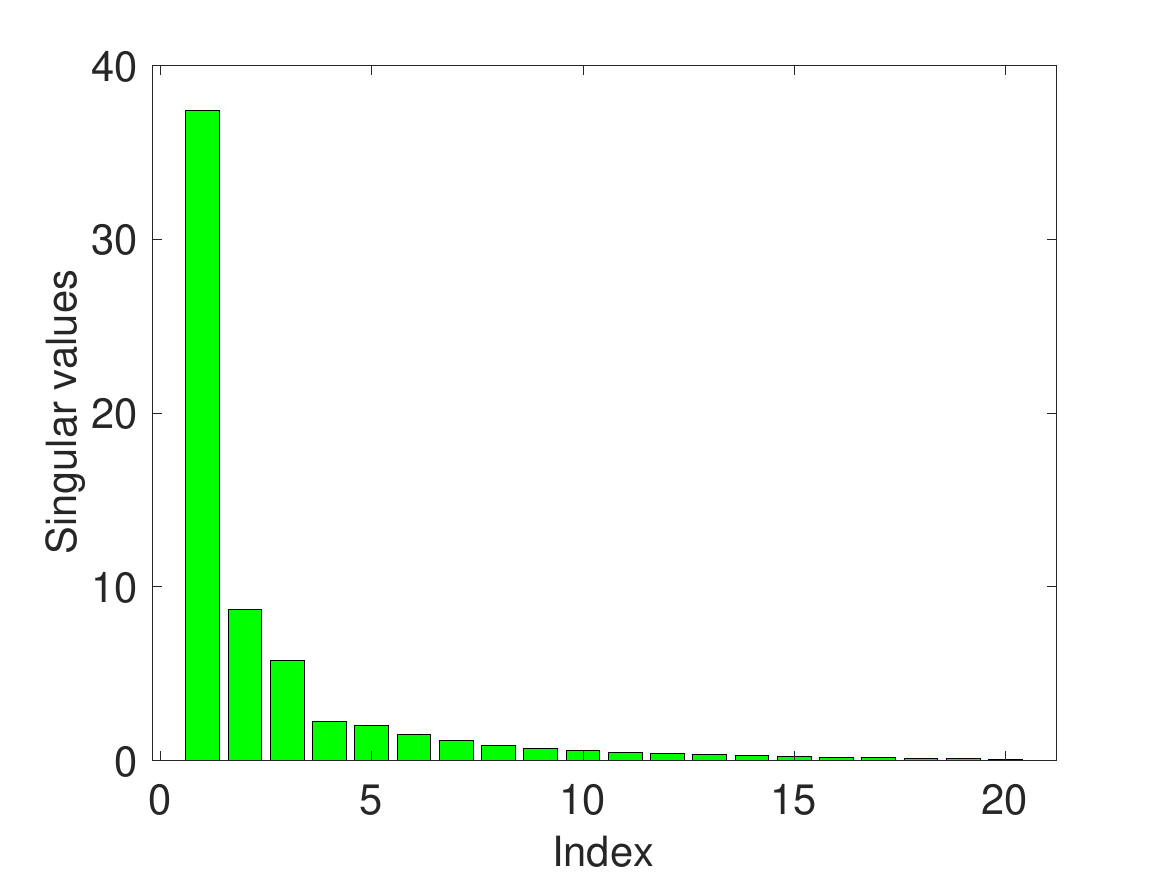} 
\caption{The spectrum of the chlorine concentration data matrix, where its singular values decay rapidly.}\label{fig:condition_number_chol}
\end{figure}

We illustrate the performance of \texttt{ScaledGD} and \texttt{ScaledGD($\lambda$)} via a real data experiment to highlight the consideration of rank selection, when the data matrix is approximately low-rank, a scenario which occurs frequently in practice. We consider a dataset\footnote{The dataset can be accessed from \url{http://www.cs.cmu.edu/afs/cs/project/spirit-1/www/}} that measures chlorine concentrations in a drinking water distribution system, over different junctions and recorded once every $5$ minutes during $15$ days. The data matrix of interest, $\bX_{\star}\in\mathbb{R}^{120\times 180}$, corresponds to the data extracted at $120$ junctions over $15$ hours. Fig.~\ref{fig:condition_number_chol} plots the spectrum of $\bX_{\star}$, where its singular values decay rapidly, suggesting it can be well approximated by a low-rank matrix.

\begin{figure}[h]
\centering
\begin{tabular}{cc}
\includegraphics[width=0.5\textwidth]{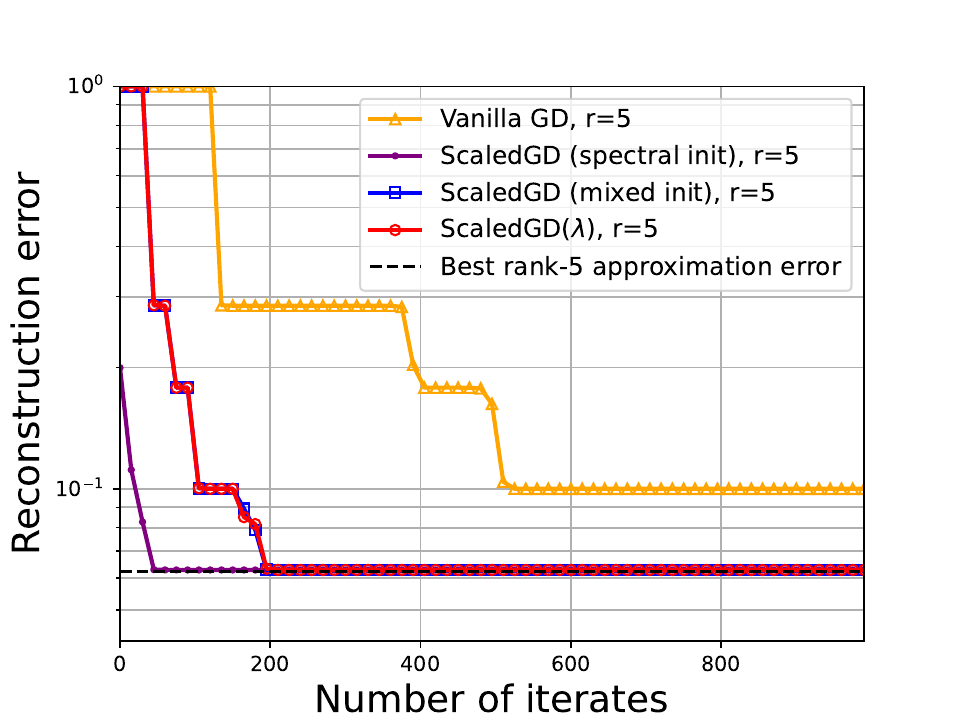}  & \includegraphics[width=0.5\textwidth]{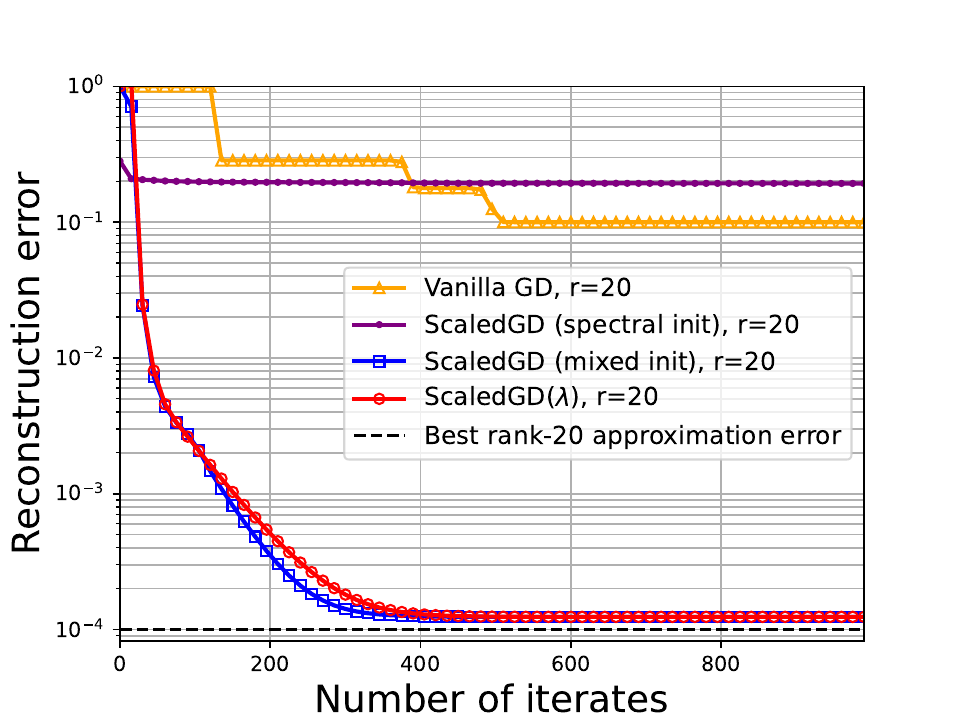} \\
(a) $r= 5$ &  (b) $r =20$
\end{tabular}
\caption{Performance comparison of different algorithms for matrix completion on the chlorine concentration dataset, 
under $r=5$ (a) and $r=20$ (b).}\label{fig:recover_chol}
\end{figure}

Since the data matrix $\bX_{\star}$ is not exactly low-rank, the choice of the rank $r$ determines how good the low-rank approximation is as well as the condition number $\kappa$: choosing a larger $r$ leads to a lower approximation error but also a higher $\kappa$. We explore the behavior of \texttt{ScaledGD} and \texttt{ScaledGD($\lambda$)} in comparison with vanilla gradient descent under different choices of $r$ in Fig.~\ref{fig:recover_chol}. Moreover, apart from the spectral initialized \texttt{ScaledGD}, we consider yet another variant of \texttt{ScaledGD} which we found useful in practice: we start with \texttt{ScaledGD($\lambda$)} for a few iterations, and switch to \texttt{ScaledGD} after it is detected $\sigma^2_{\min}(\bL_t)\gtrsim\lambda$. We call this variant \texttt{ScaledGD} with \emph{mixed initialization}. The philosophy of this variant will be introduced after we discuss the results in Fig.~\ref{fig:recover_chol}.

We consider the matrix completion setting, where we randomly observe $80\%$ of the entries in the data matrix.
Fig.~\ref{fig:recover_chol} (a) illustrates the performance of different algorithms when $r=5$.
\texttt{ScaledGD} with spectral initialization achieves the fastest convergence, while \texttt{ScaledGD($\lambda$)} and \texttt{ScaledGD} with mixed initialization take a few more iterations at the beginning to warm up. All variants of \texttt{ScaledGD} converge considerably faster than vanilla GD and approach the optimal rank-$r$ approximation error. Fig.~\ref{fig:recover_chol} (b) illustrates the performance of different algorithms when $r=20$, where the situation becomes different: \texttt{ScaledGD} with spectral initialization no longer converges, while \texttt{ScaledGD($\lambda$)} and \texttt{ScaledGD} with mixed initialization still demonstrate fast convergence to the optimal rank-$r$ approximation error. Vanilla GD, on the other hand, is still significantly slower. 

The experimental results help to explain the motivation of using \texttt{ScaledGD} with mixed initialization. The reason for the instability of spectrally initialized \texttt{ScaledGD} for large $r$ stems from the fact that spectral initialization does not cope well with overparameterization. On the other hand, small random initialization is known to help stabilize with overparameterization \cite{li2018algorithmic}, but does not integrate with \texttt{ScaledGD} since the preconditioner $({\bm L}_0^\top{\bm L}_0)^{-1}$ at the first iteration would be extremely large if the initialization $\bL_0$ were small. Therefore, initializing with \texttt{ScaledGD($\lambda$)}, which is provably robust against overparameterization and can be integrated perfectly into \texttt{ScaledGD}, becomes a  reasonable choice.

%% file: conclusion.tex
\section{Conclusions}

\label{sec:conclusion}

This chapter highlights a novel approach to provably accelerate ill-conditioned low-rank estimation via \texttt{ScaledGD}. Its fast convergence, together with low computational and memory costs by operating in the factor space, makes it a highly scalable and desirable method in practice. The performance of \texttt{ScaledGD} is also robust when the data matrix is only approximately low-rank and the observations are noisy; we refer interested readers to \cite{xu2023power} for further details.
In terms of future directions, it is of great interest to explore the design of effective preconditioners  for other statistical estimation and learning tasks, as well as further understand the implications of preconditioning in the presence of overparameterization for the asymmetric setting.